\documentclass[conference]{IEEEtran}  %
\usepackage{times}

\IEEEoverridecommandlockouts                              %

\usepackage{etex}
\usepackage[numbers,sort&compress]{natbib}
\usepackage{multicol}

\usepackage{amsmath,amsfonts,amssymb,amsthm}
\usepackage{float}
\usepackage{bm}
\usepackage{graphicx}
\usepackage{epstopdf}
\usepackage{epsfig}
\usepackage{xspace}
\usepackage{multirow}
\usepackage{hhline}
\usepackage[export]{adjustbox}
\usepackage{csvsimple}
\usepackage[dvipsnames]{xcolor}

\usepackage{graphicx}
\usepackage[colorinlistoftodos]{todonotes}
\newcommand{\journalVersion}[1]{}
\usepackage{xr}
\newcommand{\subparagraph}{} %
\usepackage{titlesec}
\graphicspath{{./figures/}}

\usepackage{tabularx}
\newcolumntype{Y}{>{\centering\arraybackslash}X} %

\usepackage{comment}
\usepackage{siunitx}
\usepackage{relsize}
\usepackage{ifthen}
\usepackage[colorinlistoftodos]{todonotes}

\usepackage[caption=false]{subfig}

\usepackage[vlined,ruled,linesnumbered]{algorithm2e}
\usepackage{graphics} %
\usepackage{rotating}
\usepackage{color}
\usepackage{enumerate}
\usepackage[T1]{fontenc}
\usepackage{psfrag}
\usepackage{epsfig} %
\usepackage{booktabs}
\usepackage{graphicx,url}
\usepackage{multirow}
\usepackage{array}
\usepackage{latexsym}
\usepackage{amsfonts}
\usepackage{amsmath}
\usepackage{amssymb}
\usepackage{xstring}
\usepackage[noend]{algorithmic}
\usepackage{multirow}
\usepackage{xcolor}
\usepackage{prettyref}
\usepackage{flexisym}
\usepackage{bigdelim}
\usepackage{breqn} %
\usepackage{listings}
\let\labelindent\relax
\usepackage{enumitem}
\usepackage{xspace}
\usepackage{bm}
\graphicspath{{./figures/}}
\usepackage{tikz}
\usetikzlibrary{matrix,calc}

\usepackage{mdwlist}

\let\itemize\undefined
\makecompactlist{itemize}{stditemize}

\newrefformat{prob}{Problem\,\ref{#1}}
\newrefformat{def}{Definition\,\ref{#1}}
\newrefformat{sec}{Section\,\ref{#1}}
\newrefformat{sub}{Section\,\ref{#1}}
\newrefformat{prop}{Proposition\,\ref{#1}}
\newrefformat{app}{Appendix\,\ref{#1}}
\newrefformat{alg}{Algorithm\,\ref{#1}}
\newrefformat{cor}{Corollary\,\ref{#1}}
\newrefformat{thm}{Theorem\,\ref{#1}}
\newrefformat{lem}{Lemma\,\ref{#1}}
\newrefformat{fig}{Fig.\,\ref{#1}}
\newrefformat{tab}{Table\,\ref{#1}}

\newcommand{\cf}{\emph{cf.}\xspace}

\newcommand{\bdmath}{\begin{dmath}}
\newcommand{\edmath}{\end{dmath}}
\newcommand{\beq}{\begin{equation}}
\newcommand{\eeq}{\end{equation}}
\newcommand{\bdm}{\begin{displaymath}}
\newcommand{\edm}{\end{displaymath}}
\newcommand{\bea}{\begin{eqnarray}}
\newcommand{\eea}{\end{eqnarray}}
\newcommand{\beal}{\beq \begin{array}{ll}}
\newcommand{\eeal}{\end{array} \eeq}
\newcommand{\beas}{\begin{eqnarray*}}
\newcommand{\eeas}{\end{eqnarray*}}
\newcommand{\ba}{\begin{array}}
\newcommand{\ea}{\end{array}}
\newcommand{\bit}{\begin{itemize}}
\newcommand{\eit}{\end{itemize}}
\newcommand{\ben}{\begin{enumerate}}
\newcommand{\een}{\end{enumerate}}

\newcommand{\calG}{{\cal G}}

\newcommand{\setal}{~\emph{et~al.}\xspace}
\newcommand{\eg}{\emph{e.g.,}\xspace}
\newcommand{\ie}{\emph{i.e.,}\xspace}
\newcommand{\myParagraph}[1]{{\bf #1.}\xspace}

\newcommand{\hide}[1]{}

\newcommand{\hiddenText}{{\color{gray} hidden text.}}
\newcommand{\hideWithText}[1]{\hiddenText}

\newcommand{\blue}[1]{{\color{blue}#1}}

\newcommand{\linkToPdf}[1]{\href{#1}{\blue{(pdf)}}}
\newcommand{\linkToPpt}[1]{\href{#1}{\blue{(ppt)}}}
\newcommand{\linkToCode}[1]{\href{#1}{\blue{(code)}}}
\newcommand{\linkToWeb}[1]{\href{#1}{\blue{(web)}}}
\newcommand{\linkToVideo}[1]{\href{#1}{\blue{(video)}}}
\newcommand{\linkToMedia}[1]{\href{#1}{\blue{(media)}}}
\newcommand{\award}[1]{\xspace} %

\newcommand{\bmat}{\left[ \begin{array}}
\newcommand{\emat}{\end{array} \right]}

\newcommand{\bal}{\begin{align}}
\newcommand{\eal}{\end{align}}

\newcommand{\TSDF}{TSDF\xspace}
\newcommand{\ESDF}{ESDF\xspace}
\newcommand{\GVD}{GVD\xspace}
\newcommand{\ESDFs}{ESDFs\xspace}
\newcommand{\name}{Hydra\xspace}

\newcommand{\igx}{\emph{GT-Trajectory}\xspace}

\newcommand{\ivl}{\emph{VIO+V-LC}\xspace}
\newcommand{\ivd}{\emph{VIO+SG-LC}\xspace}

\newcommand{\percFound}{\emph{\% Found}\xspace}
\newcommand{\percCorrect}{\emph{\% Correct}\xspace}
\newcommand{\positionError}{\emph{Position Error}\xspace}

\newcommand{\SPIN}{Spatial Perception System\xspace}

\usepackage[colorlinks=true, allcolors=blue, bookmarks=true]{hyperref}

\titlespacing*{\section}{0pt}{4mm}{2mm}
\titlespacing*{\subsection}{0pt}{2mm}{2mm}

\newcommand{\isArxiv}[2]{#1} %

\newcommand{\tc}[1]{\textcolor{OliveGreen}{#1}\xspace}
\newcommand{\lc}[1]{\textcolor{magenta}{#1}\xspace}
\renewcommand{\tc}[1]{#1}
\renewcommand{\lc}[1]{#1}
\newcommand{\hydraURL}{\url{https://github.com/MIT-SPARK/Hydra}\xspace}

\newcommand{\acceptance}{%
\begin{tikzpicture}[overlay, remember picture]
\path (current page.north east) ++(-3.2,-0.4) node[below left] {
This paper has been accepted for publication at the 2022 Robotics: Science and Systems Conference.
};
\end{tikzpicture}
\begin{tikzpicture}[overlay, remember picture]
\path (current page.north east) ++(-6.5,-0.8) node[below left] {
Please cite the paper as: N. Hughes, Y. Chang, and L. Carlone,
};
\end{tikzpicture}
\begin{tikzpicture}[overlay, remember picture]
\path (current page.north east) ++(0.0,-1.2) node[below left] {
``Hydra: A Real-time Spatial Perception System for 3D Scene Graph Construction and Optimization,'' \emph{Robotics: Science and Systems (RSS)}, 2022.
};
\end{tikzpicture}
}

\begin{document}

\title{\name: A Real-time \tc{\SPIN} for \\ 3D Scene Graph Construction and Optimization}

\author{
    \authorblockN{Nathan Hughes, Yun Chang, Luca Carlone}
    \authorblockA{Laboratory for Information \& Decision Systems (LIDS)\\
                  Massachusetts Institute of Technology\\
                  Cambridge, USA\\
                  Email: \{na26933, yunchang, lcarlone\}@mit.edu}
}

\maketitle
\isArxiv{\acceptance}{}

\begin{abstract}
3D scene graphs have recently emerged as a powerful high-level representation of 3D environments.
A 3D scene graph describes the environment as a layered graph where nodes represent spatial concepts at multiple levels of abstraction
(from low-level geometry to high-level semantics including objects, places, rooms, buildings, etc.)
and edges represent relations between concepts. While 3D scene graphs can serve as an advanced ``mental model'' for robots,
how to build such a rich representation in real-time is still uncharted territory.

This paper describes %
\lc{a \emph{real-time} \SPIN}, a
suite of algorithms to build a 3D scene graph from sensor data in real-time.
Our first contribution is to develop real-time algorithms to incrementally construct the layers of a scene graph as the robot explores the environment;
these algorithms build a local Euclidean Signed Distance Function (\ESDF) around the current
robot location, extract a topological map of places from the ESDF, and
then segment the places into rooms using an approach inspired by community-detection techniques.
Our second contribution is to investigate loop closure detection and optimization in 3D scene graphs.
We show that 3D scene graphs allow defining \emph{hierarchical descriptors} for loop closure detection;
 our descriptors capture statistics across layers in the scene graph, ranging from low-level visual appearance to summary statistics about objects and places.
 We then propose the first algorithm to optimize a 3D scene graph in response to loop closures;
 our approach relies on \emph{embedded deformation graphs} to simultaneously correct all layers of the scene graph.
    We implement the proposed \lc{Spatial Perception System} into a highly parallelized architecture, named \emph{\name}\footnote{\tc{\name is available at \hydraURL}}, that
combines fast early and mid-level perception processes (\eg local mapping)
with slower high-level perception (\eg global optimization of the scene graph).
We evaluate \name on simulated and real data and show it is able to reconstruct 3D scene graphs with an
accuracy comparable with batch offline methods despite running online.

\end{abstract}


\section{Introduction} %
\label{sec:introduction}
The next generation of robots and autonomous systems will be required to build 
persistent high-level representations of unknown environments in real-time. 
\emph{High-level} representations are required for a robot to understand and execute instructions from humans 
 (\eg ``bring me the cup of tea I left on the dining room table''); high-level representations also enable 
 fast planning (\eg by allowing planning over compact abstractions rather than dense low-level geometry).
Such representations must be built in \emph{real-time} to support just-in-time decision-making.
Moreover, these representations must be \emph{persistent} to support long-term autonomy: 
(i) they need to scale to large environments, (ii) they should allow for corrections as new evidence is collected by the robot, and (iii) their size should only grow with the size of the environment they model.

\begin{figure}
    \centering
    \includegraphics[trim={36, 0, 140, 0}, clip, width=0.99\columnwidth]{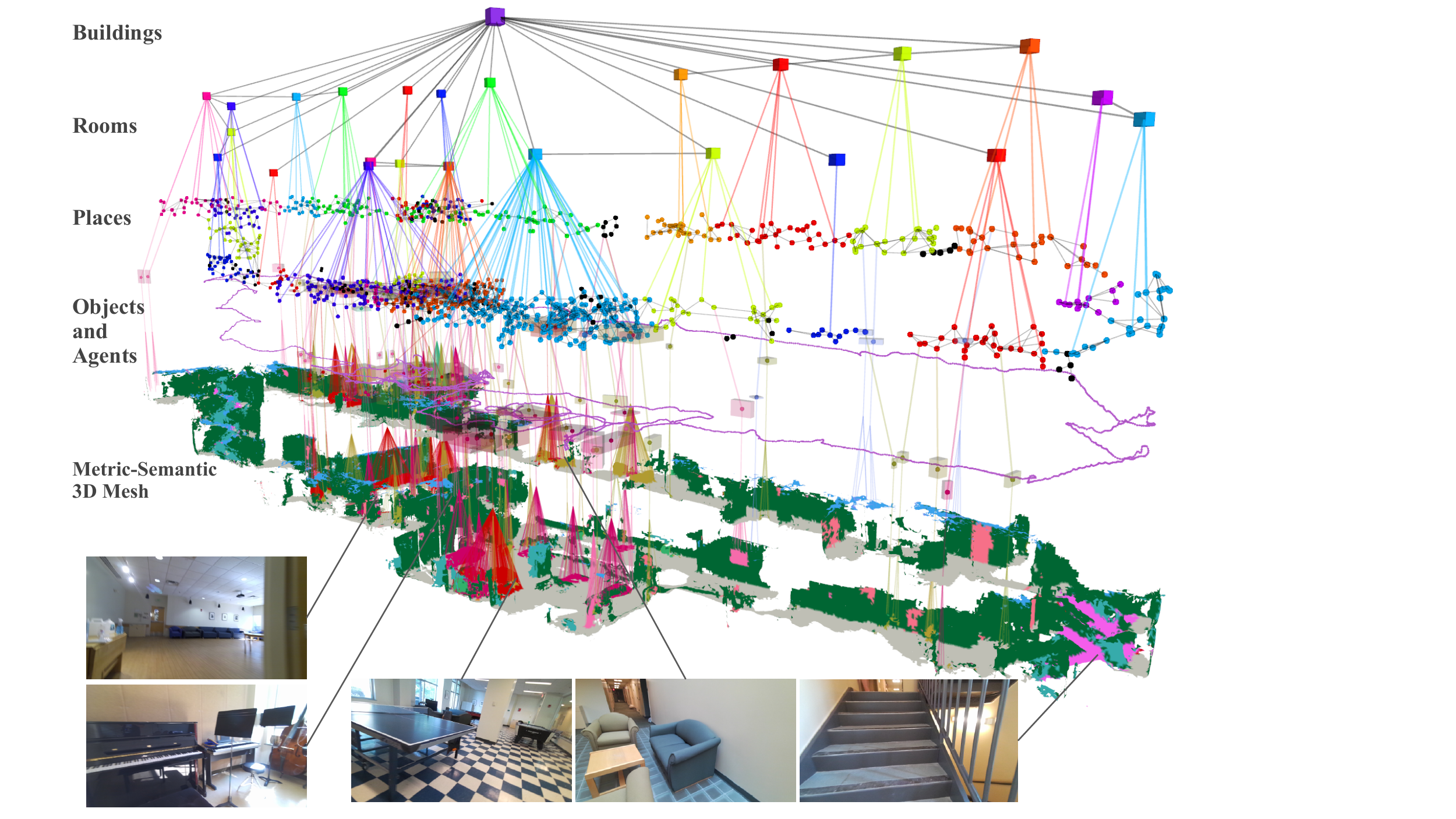} \vspace{-5mm}
    \caption{We present \emph{\name}, a highly parallelized architecture to build 3D scene graphs
    from sensor data in real-time. The figure shows sample input data and the 3D scene graph created by \name 
    in a large-scale real environment.} \vspace{-5mm} \label{fig:intro}
\end{figure}
 
3D Scene Graphs~\cite{Rosinol21ijrr-Kimera,Armeni19iccv-3DsceneGraphs,Rosinol20rss-dynamicSceneGraphs,Kim19tc-3DsceneGraphs,Wald20cvpr-semanticSceneGraphs,Wu21cvpr-SceneGraphFusion} 
have recently emerged as powerful high-level representations of 3D environments. A 3D scene graph (Fig.~\ref{fig:intro} \lc{and Fig.~\ref{fig:example_dsg}}) is a layered graph where nodes represent spatial concepts at multiple levels of abstraction (from low-level geometry to objects, places, rooms, buildings, etc.) and edges represent relations between concepts. 
 Armeni\setal~\cite{Armeni19iccv-3DsceneGraphs} pioneered the use of 3D scene graphs in computer vision and 
 proposed the first algorithms to parse a metric-semantic 3D mesh 
 into a 3D scene graph. Kim\setal~\cite{Kim19tc-3DsceneGraphs} reconstruct a 3D scene graph of objects and their relations. 
 Rosinol\setal~\cite{Rosinol21ijrr-Kimera,Rosinol20rss-dynamicSceneGraphs} propose a novel 3D scene graph model that 
 (i) is built directly from sensor data,
 (ii) includes a subgraph of places (useful for robot navigation), 
 (iii) models objects, rooms, and buildings, and 
 (iv) captures moving entities in the environment.
 More recent work~\cite{Wald20cvpr-semanticSceneGraphs, Wu21cvpr-SceneGraphFusion,Izatt21tr-sceneGraphs,Gothoskar21arxiv-3dp3} infers objects and relations from point clouds, RGB-D sequences, or object detections.

While 3D scene graphs can serve as an advanced ``mental model'' for robots,
how to build such a rich representation in real-time remains uncharted territory.
The works~\cite{Kim19tc-3DsceneGraphs,Wald20cvpr-semanticSceneGraphs, Wu21cvpr-SceneGraphFusion} allow real-time operation but are restricted to ``flat'' 3D scene graphs and are mostly concerned with objects and their relations while disregarding the top layers in Fig.~\ref{fig:intro}.
The works~\cite{Rosinol21ijrr-Kimera,Armeni19iccv-3DsceneGraphs,Rosinol20rss-dynamicSceneGraphs}, which focus on building truly hierarchical representations, run offline and require several minutes to build a 3D scene graph (\cite{Armeni19iccv-3DsceneGraphs} even assumes the availability of a correct and complete metric-semantic mesh of the environment built beforehand).
Extending \tc{our prior} works~\cite{Rosinol21ijrr-Kimera,Rosinol20rss-dynamicSceneGraphs} to operate in real-time is non-trivial.
These works utilize an Euclidean Signed Distance Function (\ESDF) of the entire environment to build the 3D scene graph. %
Unfortunately, \tc{the memory required for} \ESDFs scale poorly in the size of the environment~\cite{Oleynikova17iros-voxblox}.
Moreover, the extraction of places and rooms in~\cite{Rosinol21ijrr-Kimera,Rosinol20rss-dynamicSceneGraphs} involves batch algorithms that process the entire \ESDF, whose computational cost grows over time and is incompatible with real-time operation. 
Finally, the \ESDF is reconstructed from the robot trajectory estimate
\tc{which keeps changing in response to loop closures. The approaches in~\cite{Rosinol21ijrr-Kimera,Rosinol20rss-dynamicSceneGraphs} would therefore need to rebuild the scene graph from scratch after every loop closure, 
clashing with real-time operation.}

\lc{The main} \tc{motivation} \lc{of this paper is to overcome these challenges and develop 
the first \emph{real-time} {\SPIN}, a 
suite of algorithms and implementations to build a} \tc{hierarchical 3D scene graph, such as the one pictured in Fig.~\ref{fig:intro}}, \lc{from sensor data in real-time.}

Our first contribution is to develop real-time algorithms to incrementally reconstruct the layers of a scene graph as the robot explores the environment. 
The proposed algorithms reconstruct a \emph{local} \ESDF of the robot's surroundings and incrementally convert the \ESDF into a metric-semantic 3D mesh as well as a \emph{Generalized Voronoi Diagram}, from which a topological graph of places can be quickly extracted. This computation is incremental and runs in constant-time regardless of the size of the environment. 
Our algorithms also perform a fast and scalable room segmentation \tc{using} a community-detection inspired approach that
clusters the graph of places into rooms in a matter of milliseconds.

Our second contribution is to investigate loop closure detection and optimization in 3D scene graphs.
We propose a novel hierarchical approach for loop closure detection: the proposed approach involves (i)~a \emph{top-down 
loop closure detection} that uses hierarchical descriptors ---capturing statistics across layers in the scene graph--- 
to find putative loop closures
and (ii)~a \emph{bottom-up geometric verification} that attempts estimating the loop closure pose 
 by registering putative matches. %
 Then, we propose the first algorithm to optimize a 3D scene graph in response to loop closures; 
 our approach relies on \emph{embedded deformation graphs} to simultaneously correct all layers of the scene graph, \ie the 3D mesh, places, objects, and rooms.

Our final contribution is to develop a real-time architecture and implementation, and demonstrate the resulting 
\lc{Spatial Perception System} 
on challenging simulated and real data. 
In particular, we propose a highly parallelized implementation, named \emph{\name}, that 
combines fast early and mid-level perception processes (\eg local mapping) 
with slower high-level perception (\eg global optimization of the scene graph).
We evaluate \name in several heterogeneous environments, including an apartment complex, an office building, and a subway. Our experiments show that (i) we can reconstruct 3D scene graphs of large, real environments in real-time,
 (ii) our online algorithms achieve an accuracy comparable 
to batch offline methods, 
and (iii) our loop closure detection approach outperforms standard approaches based on bag-of-words and visual-feature matching in terms of quality and quantity of detected loop closures. 
\lc{The source code of \name is publicly available at \hydraURL.} 

\section{Related Work} %

\myParagraph{Metric-semantic and Hierarchical Mapping}
The last few years have seen a surge of interest towards \emph{metric-semantic mapping}, simultaneously triggered by the maturity of traditional 3D reconstruction and SLAM techniques, and by the novel opportunities for semantic understanding afforded by deep learning. The literature has focused on both object-based maps~\cite{Salas-Moreno13cvpr,Dong17cvpr-XVIO,Mo19iros-orcVIO,Nicholson18ral-quadricSLAM,Bowman17icra,Ok21icra-home} and dense maps, including volumetric models~\cite{McCormac17icra-semanticFusion,Grinvald19ral-voxbloxpp,Narita19iros-metricSemantic}, point clouds~\cite{Behley19iccv-semanticKitti,Tateno15iros-metricSemantic,Lianos18eccv-VSO}, 
and 3D meshes~\cite{Rosinol20icra-Kimera,Rosu19ijcv-semanticMesh}. Some approaches combine objects and dense map models~\cite{Li16iros-metricSemantic,McCormac183dv-fusion++,Xu19icra-midFusion,Schmid21arxiv-panoptic}.
These approaches are not concerned with estimating higher-level semantics (\eg rooms) and typically return 
 dense models that might not be directly amenable for navigation~\cite{Oleynikova18iros-topoMap}. 

A second research line focuses on building \emph{hierarchical map} models.
Hierarchical maps have been pervasive in robotics since its inception~\cite{Kuipers00ai,Kuipers78cs,Chatila85,Thrun02a}. Early work focuses on 2D maps and investigates the use of hierarchical maps to resolve the apparent divide between 
metric and topological representations~\cite{Ruiz-Sarmiento17kbs-multiversalMaps,Galindo05iros-multiHierarchicalMaps,Zender08ras-spatialRepresentations}.   
More recently, 3D scene graphs have been proposed as expressive hierarchical models for 3D environments.
 Armeni\setal~\cite{Armeni19iccv-3DsceneGraphs} model the environment as a graph including low-level geometry (\ie a metric-semantic mesh), objects, rooms, and camera locations. Rosinol\setal~\cite{Rosinol21ijrr-Kimera,Rosinol20rss-dynamicSceneGraphs} augment the model with a topological map of places, as well as a layer describing dynamic entities in the environment. 
 The approaches in~\cite{Armeni19iccv-3DsceneGraphs,Rosinol21ijrr-Kimera,Rosinol20rss-dynamicSceneGraphs} are designed for offline \lc{use}.
 Other papers focus on reconstructing a graph of objects and their relations~\cite{Kim19tc-3DsceneGraphs,Wald20cvpr-semanticSceneGraphs,Wu21cvpr-SceneGraphFusion}. Wu\setal~\cite{Wu21cvpr-SceneGraphFusion} 
 predict objects and relations in real-time using a \lc{graph neural network}. 
Izatt and Tedrake~\cite{Izatt21tr-sceneGraphs} parse objects and relations into a scene grammar model
using mixed-integer programming. Gothoskar\setal~\cite{Gothoskar21arxiv-3dp3} use an MCMC approach.

A somewhat parallel research line investigates how to \emph{parse the layout of a building} from 2D or 3D data.
A large body of work focuses on parsing 2D maps~\cite{Bormann16icra-roomSegmentationSurvey}, 
including rule-based~\cite{Kleiner17iros-roombaRoomSegmentation} and learning-based methods~\cite{Liu18eccv-floorNet}.
Friedman\setal~\cite{Friedman07ijcai-voronoiRF} compute a Voronoi graph from a 2D occupancy grid, which is then labeled using a conditional random field. 
Recent work focuses on 3D data. Liu\setal~\cite{Liu18eccv-floorNet} and Stekovic\setal~\cite{Stekovic21arxiv-monteFloor}
project 3D point clouds to 2D maps, which however is not directly applicable to multi-story buildings. 
Furukawa\setal~\cite{Furukawa09iccv} reconstruct floor plans from images using multi-view stereo combined with a Manhattan World assumption. Lukierski\setal~\cite{Lukierski17icra-floorPlan} use dense stereo from an omni-directional camera to
fit cuboids to objects and rooms.   
Zheng\setal~\cite{Zheng20pami-buildingFusion} detects rooms by performing region growing on a 3D metric-semantic model.

\myParagraph{Loop Closures Detection and Optimization}
Established approaches for visual loop closure detection in robotics trace back to  place recognition and image retrieval techniques in computer vision; these approaches are broadly adopted in SLAM pipelines but are known to suffer from 
appearance and viewpoint changes~\cite{Lowry16tro-surveyPlaceRecognition}.
Recent approaches investigate place recognition %
using image sequences~\cite{Schubert21rss-fastICM,Garg21ral-seqnet} 
or deep learning~\cite{Arandjelovic16cvpr-netvlad}. 
More related to our proposal is the set of papers leveraging %
 semantic information for loop closure detection.
Gawel\setal~\cite{Gawel18ral-xview} perform object-graph-based loop closure detection using random-walk descriptors built from 2D images. 
Liu\setal~\cite{Liu19icra-globalLocalizationObjects} use similar object-based descriptors but built from a 3D reconstruction.
Lin\setal~\cite{Lin21ral-topologyAwareObjectLocalization} adopt random-walk object-based descriptors 
and then compute loop closure poses via object registration.  
Qin\setal~\cite{Qin21jvcir-semantic} propose an object-based approach based on subgraph similarity matching.
Zheng\setal~\cite{Zheng20pami-buildingFusion} propose a room-level loop closure detector.

After a loop closure is detected, the map needs to be corrected accordingly. 
While this process is easy in sparse (\eg landmark-based) representations~\cite{Cadena16tro-SLAMsurvey}, it 
is non-trivial to perform in real-time when using dense representations. 
St\"{u}ckler and Behnke~\cite{Stuckler14jvcir} and Whelan\setal~\cite{Whelan16ijrr-elasticFusion} 
optimize a map of \emph{surfels}, 
to circumvent the need to correct structured representations (\eg meshes or voxels).
Dai\setal~\cite{Dai17tog-bundlefusion} propose reintegrating a volumetric map after each loop closure. 
Reijgwart\setal~\cite{Reijgwart20ral-voxgraph} correct drift in volumetric representations by breaking the 
map into submaps that can be rigidly re-aligned after loop closures.
Whelan\setal~\cite{Whelan15ijrr} propose a 2-step optimization that first corrects the robot trajectory 
 and then deforms the map (represented as a point cloud or a mesh) using a deformation graph approach~\cite{Sumner07siggraph-embeddedDeformation}.
Rosinol\setal~\cite{Rosinol21ijrr-Kimera} \lc{unify} the two steps into a single pose graph and mesh optimization.
None of these works is concerned with simultaneously correcting multiple hierarchical representations.

\section{Real-time Incremental \\ 3D Scene Graph Layers Construction} %
\label{sec:incrementalLayers}

This section describes how to construct the layers of a 3D scene graph given an odometric estimate of the
robot trajectory (\eg from visual-inertial odometry). Then, Section~\ref{sec:LCD-and-SGO} discusses how to correct the graph in response to loop closures.

\begin{figure}
    \centering
    \includegraphics[trim={65, 10, 160, 0}, clip, width=\columnwidth]{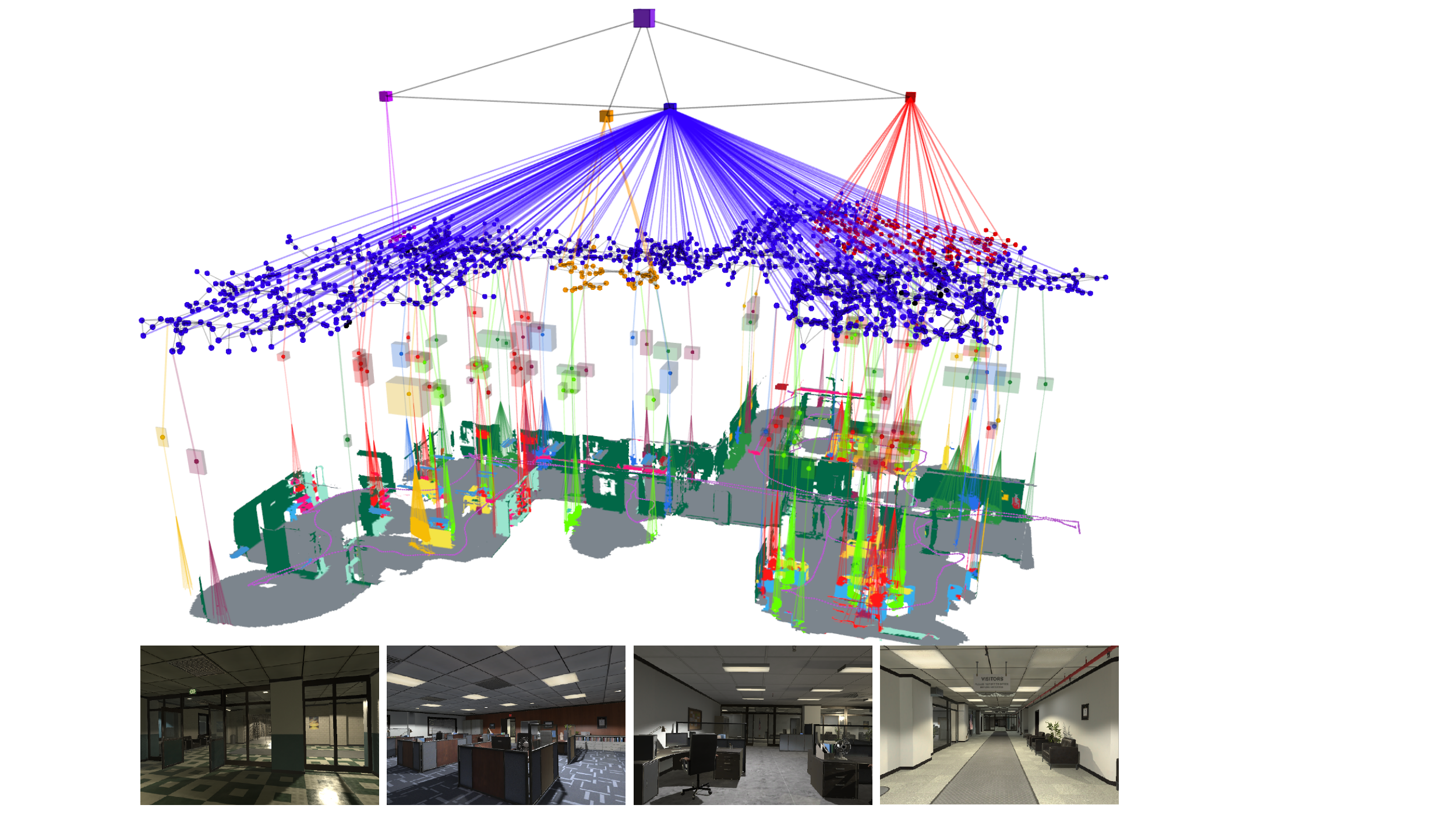} \vspace{-7mm}
    \caption{3D scene graph created by \name in the uH2 Office dataset. }\vspace{-7mm} \label{fig:example_dsg}
\end{figure}

We focus on indoor environments and adopt the 3D scene graph model introduced in~\cite{Rosinol20rss-dynamicSceneGraphs} and visualized in Fig.~\ref{fig:intro} \lc{and Fig.~\ref{fig:example_dsg}}.
In this model,
\emph{Layer~1} is a metric-semantic 3D mesh.
\emph{Layer~2} is a subgraph of objects and agents; each object has a semantic label, a centroid, and a bounding box,
while each agent is modeled by a pose graph describing its trajectory (in our case the robot itself is the only agent).
\emph{Layer~3} is a subgraph of \emph{places} (essentially, a topological map) where
each place is an obstacle-free location and an edge between places denotes straight-line traversability.
\emph{Layer~4} is a subgraph of rooms where each room has a centroid, and edges connect adjacent rooms.
\emph{Layer~5} is a building node connected to all rooms (we assume the robot maps a single building). Edges connect nodes within each layer (\eg to model traversability between places or rooms)
or across layers (\eg to model that mesh vertices belong to an object, that an object is in a certain room, \lc{or that a room belongs to a building}).

Next, we present an
approach to construct Layers 1-3 (Section~\ref{sec:mesh_objects_places}) and
to segment places into rooms (Section~\ref{sec:rooms}).

\begin{figure}
    \hspace{-5mm}
    \subfloat[GVD and 3D Mesh]{
        \centering
        \includegraphics[width=0.56\columnwidth]{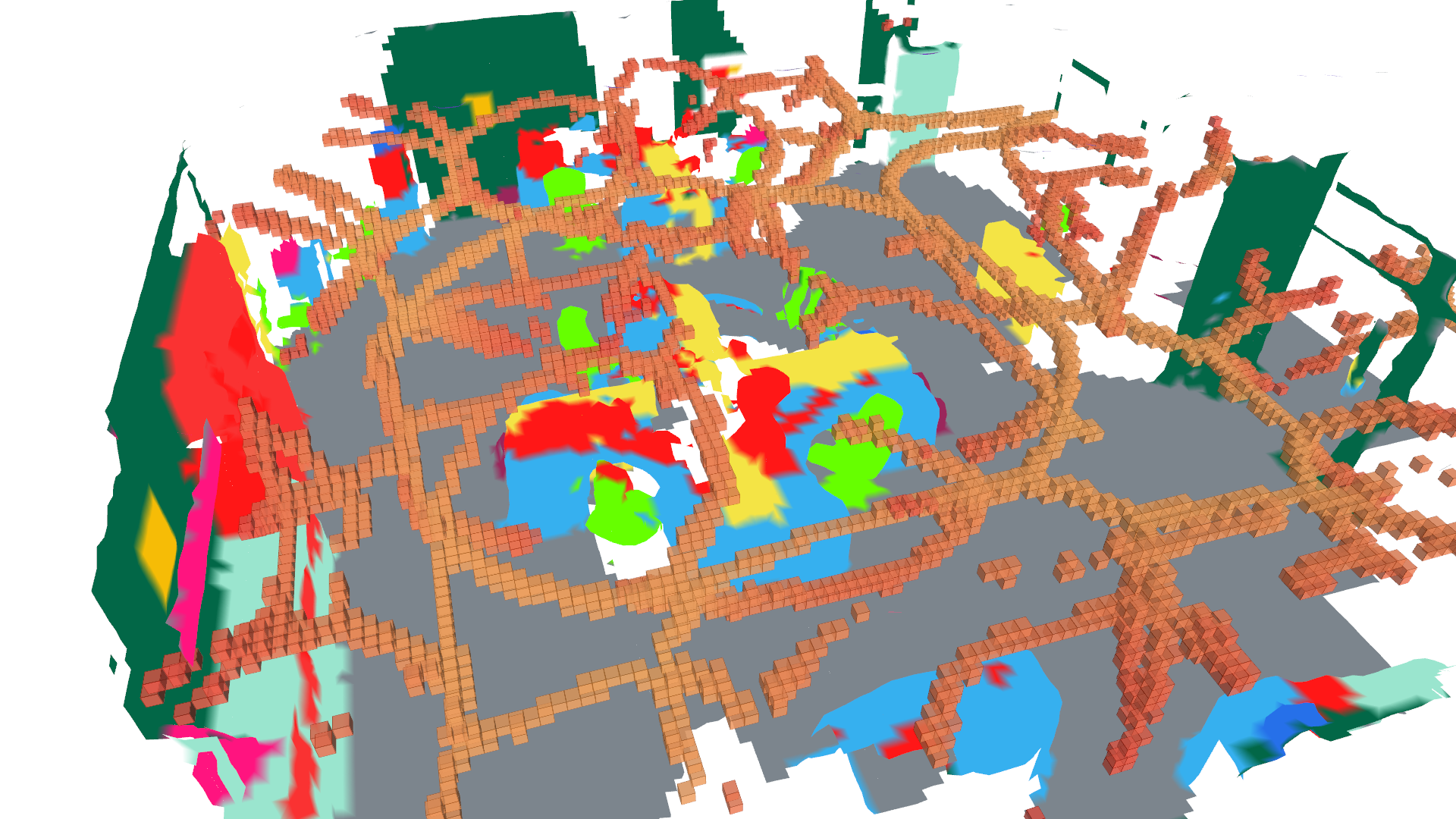}\label{fig:gvd}
    }
    \subfloat[Room Detection]{
        \centering
        \includegraphics[width=0.43\columnwidth]{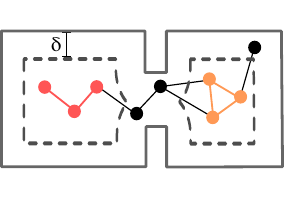}\label{fig:rooms}
    }
    \caption{(a) GVD (orange blocks) and mesh inside the active window. %
             (b) Room detection: connected components (in orange and red) in the subgraph of places induced by a dilation distance $\delta$ (walls are shown in gray, dilated walls are dashed, places that disappear after dilation are in black).}\vspace{-5mm}
\end{figure}

\subsection{Layers 1-3: Mesh, Objects, and Places}
\label{sec:mesh_objects_places}

\myParagraph{Mesh and Objects}
The real-time construction of the metric-semantic 3D mesh \tc{(\emph{Layer 1} of the 3D scene graph)} is an extension  \lc{of Kimera~\cite{Rosinol21ijrr-Kimera}}, with small but important modifications.
\tc{Kimera~\cite{Rosinol21ijrr-Kimera}} uses Voxblox~\cite{Oleynikova17iros-voxblox} to
integrate semantically-labeled point clouds into a
 Truncated Signed Distance Field (TSDF)
and an \ESDF of the environment, while also performing Bayesian inference over the semantic label of each voxel. 

Contrarily to \tc{Kimera~\cite{Rosinol21ijrr-Kimera}} and building on the implementation in~\cite{Oleynikova17iros-voxblox}, we {spatially window} the \TSDF and \ESDF
and only form a volumetric model of the robot's surroundings within a \lc{user-specified} radius (8m in our implementation);
the radius is chosen to bound the amount of memory used by the \ESDF.
Within this ``active window'' we extract the 3D metric-semantic mesh using Voxblox' marching cubes implementation
and the places (as described in the next paragraph); as the mesh and places move outside the active window, they are passed to the Scene Graph Frontend (Section~\ref{sec:LCD-and-SGO}).
We also modify the marching cubes algorithm to label the \TSDF voxels that correspond to zero-crossings
(\ie~the voxels containing a surface); we call these voxels ``parents'' and keep track of the corresponding mesh vertices. %
Then, for each \ESDF voxel %
--which already stores a distance to the closest obstacle--
 we additionally keep track of which parent is closest to the voxel.
When extracting the places from the \ESDF we use the parents to associate each place to the closest vertex in the 3D mesh.
After extracting the 3D mesh within the active window,
we segment objects \tc{(\emph{Layer 2} of the 3D scene graph}) by performing Euclidean clustering
of the 3D metric-semantic mesh vertices; in particular, we independently cluster vertices belonging to each semantic class.
As in \lc{Kimera~\cite{Rosinol21ijrr-Kimera}, the result of the Euclidean clustering is then used to estimate a centroid and bounding box for each putative object. 
During incremental operation, if a putative object overlaps\footnote{\lc{In practice, we consider two objects overlapping if the centroid of one object is contained in the other object's bounding box, a proxy for spatial overlap measures such as Intersection over Union (IoU). 
}}
 with an existing object node of the same semantic class in the scene graph, 
 we merge them together by adding new mesh vertices to the previous object node;
 if the new object does not correspond to an existing object node it is added as a new node.}

\myParagraph{Places}
\lc{Kimera~\cite{Rosinol21ijrr-Kimera}}
builds a monolithic \ESDF of the environment and then
uses~\cite{Oleynikova18iros-topoMap} to extract
the places subgraph \tc{(\emph{Layer 3} in the 3D scene graph)}.
We instead implement an approach that incrementally
extracts the subgraph of places
using a
Generalized Voronoi Diagram (\GVD, shown in Figure~\ref{fig:gvd}) \lc{built on the fly during the \ESDF integration}.  %
The GVD is the set of voxels that are equidistant to at least 2 obstacles (``basis points'' \tc{or ``parents''}),
and intuitively forms a skeleton of the environment~\cite{Oleynikova18iros-topoMap}.
We obtain the GVD as a byproduct of the \ESDF integration inside the active window, following the approach in~\cite{Lau13ras-efficientGridRepresentations}. In particular, the voxels belonging to the GVD can be easily detected from the wavefronts of the brushfire algorithm used to update the \ESDF.

After the GVD of the active window is computed, we \tc{incrementally sparsify the GVD into a subgraph of places by modifying}
 the batch approach of~\cite{Oleynikova18iros-topoMap}. \lc{Intuitively, we select a subset of GVD {voxels} to become places nodes and connect them with edges to form the graph of places.}
\tc{After each update to the GVD, we iterate through each new voxel member of the GVD with enough basis points
 \lc{(3 in our implementation)} to \lc{create} a node or edge. Voxels are considered nodes
if} the new GVD voxel either has enough basis points (\lc{\ie $\geq 4$}) or if the neighborhood of the voxel matches a template proposed
by~\cite{Oleynikova18iros-topoMap} to identify corner voxels\tc{.}
To identify edges between nodes, we alternate between two phases.
First, we label GVD voxels with the nearest node ID via flood-fill, starting from the labels produced from the previous \ESDF integration
\tc{to generate} a putative set of edges from all neighboring node IDs.
\tc{As a second phase, we split putative edges where the straight-line edge deviates too far from the GVD voxels connecting the two nodes by inserting a new node} \tc{at the GVD voxel with the maximum deviation.}
During the \tc{first phase (flood-fill)}, we also merge nearby nodes.
After a fixed number of iterations \tc{of the two phases}, we
then add the identified edges to the sparse graph, and remove any disconnected nodes.
Finally, we add extra edges between disconnected components inside the active window to make the subgraph of places connected.
\subsection{Layer 4: Room Detection}
\label{sec:rooms}

The room detection approach in \tc{Kimera~\cite{Rosinol21ijrr-Kimera}} requires
a volumetric representation of the entire environment
and makes assumptions on the room geometry (\eg ceiling height) that do not easily extend to arbitrary (and possibly multi-story) buildings.
To resolve these issues,
we present a novel approach \tc{for constructing \emph{Layer 4} of the 3D scene graph}
that segments rooms directly from the sparse subgraph of places.
The subgraph of places, that we denote as $\calG_p$,
can be the one produced by the approach in Section~\ref{sec:mesh_objects_places},
or the optimized one computed after loop closures, as described in Section~\ref{sec:LCD-and-SGO}.

Our approach is based on two key insights.
The first is that dilation operations on the voxel-based map help expose rooms in the environment:
if we inflate obstacles, small apertures in the environment (\ie doors) will gradually close, naturally partitioning the
voxel-based map into disconnected components (\ie rooms).
The second insight is that each node in our place subgraph $\calG_p$ stores a distance to its
closest obstacle (Section~\ref{sec:mesh_objects_places});
therefore, dilation operations in the voxel-based map can be directly mapped into
topological changes in $\calG_p$. More precisely, if we dilate the map by a distance $\delta$, every
place with obstacle distance smaller than $\delta$ will disappear from the graph (since it will
no longer be in the free space). A visualization of this idea is given in Fig.~\ref{fig:rooms}.

These insights motivate our approach for room detection.
We dilate the map by increasing distances $\delta$ (\eg 10 distances uniformly spaced in {$[0.45,1.2]$}m).
For each dilation distance, we prune the subgraph of places by discarding nodes with distance smaller than $\delta$ (and their edges); we call the pruned subgraph $\calG_{p,\delta}$.
We count the number of connected components in $\calG_{p,\delta}$ (intuitively, for a suitable choice of $\delta$, the connected components will correspond to the rooms in the environment).\footnote{In practice, we combine the dilation and the connected component computation, \ie~we never explicitly compute the dilated subgraph and instead just prevent nodes and
edges from being visited during the breadth-first search for connected
components if they fall below the distance threshold $\delta$.} Then we compute the median number of connected components $n_r$ (to gain robustness to the choice of $\delta$)
and select the largest $\calG_{p,\delta^\star}$ that has $n_r$ connected components.
Finally, since $\calG_{p,\delta^\star}$ might miss some of the nodes in the original graph $\calG_{p}$,
we assign these unlabeled nodes via a \lc{partially seeded} clustering technique.
In particular, we
use a greedy modularity-based community detection approach
from~\cite{Blondel08jsm-Louvain} that involves iteratively attempting to assign
each node in the graph to a community (\ie a room) that would result in the largest
increase in modularity. We seed the initial communities to be the detected
connected components in $\calG_{p,\delta^\star}$ and only iterate through unlabeled nodes.
This both produces qualitatively consistent results and scales significantly better than related techniques (\eg
spectral clustering is a popular method for clustering graphs but requires a more expensive eigen-decomposition of the Laplacian of $\calG_{p}$).

\lc{Our novel room detection method provides two advantages.
First, reasoning over 3D free-space instead of a 2D occupancy grid (\eg~\cite{Kleiner17iros-roombaRoomSegmentation}) reduces the impact of clutter in the environment
 Second, the parameters for the method can be set to work for a variety of environments (\ie the two main parameters are the minimum and maximum opening size between rooms).  At the same time, our room detection only reasons over the topology of the environment; as such it will fail to segment semantically distinct rooms in an open floor-plan.}

\section{Persistent Representations: Loop Closure Detection and 3D Scene Graph Optimization}
\label{sec:LCD-and-SGO}

\begin{figure}[t]
    \centering
    \includegraphics[width=0.95\columnwidth]{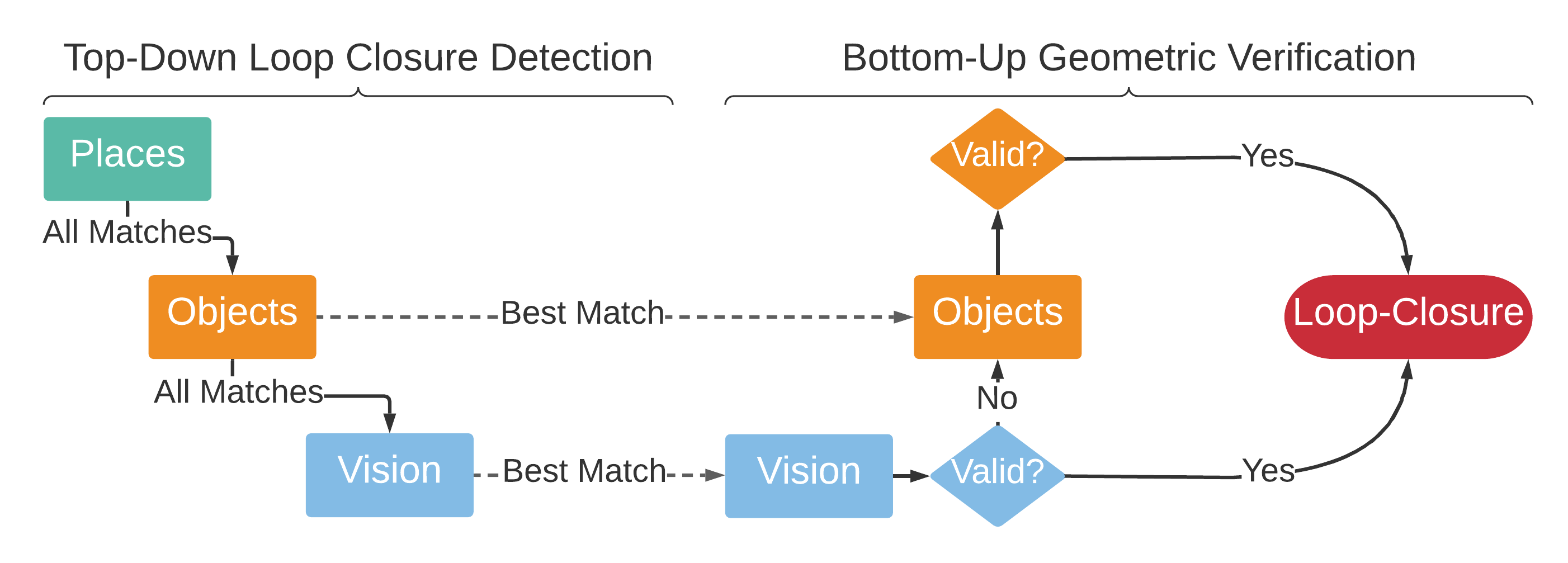}
    \vspace{-3mm}
    \caption{
    Loop closure detection (left) and geometric verification (right).
    To find a match, we ``descend'' the 3D scene graph layers, comparing descriptors.
    We then ``ascend'' the 3D scene graph layers, attempting registration.}
    \vspace{-6mm} \label{fig:lcd}
\end{figure}
 
While the previous section describes how to incrementally build the layers of an
``odometric'' 3D scene graph (the layers are built from current odometry estimates), this section describes how to detect loop closures (Section~\ref{sec:LCD}) and
how to correct the
scene graph after loop closures (Section~\ref{sec:SGO}).
\subsection{Loop Closure Detection and Geometric Verification}
\label{sec:LCD}

We augment visual loop closure detection
and geometric verification by
using multiple layers in the scene graph.

\myParagraph{Top-down Loop Closure Detection}
The agent layer in the 3D scene graph stores a pose graph describing the robot trajectory; we refer to these poses as the \emph{agent nodes}. In our implementation, we store a keyframe for each agent node, from which appearance information can be extracted.
Loop closure detection then aims at finding a past agent node that matches (\ie observed the same portion of the scene seen by) the latest agent node (corresponding to the current robot pose).

For each agent node, we construct a hierarchy of descriptors describing statistics of the node's surroundings,
from low-level appearance to objects semantics and places geometry.
 The descriptors are computed only once, when a new agent node is instantiated.
At the lowest level, our hierarchical descriptors include
 standard DBoW2 %
  appearance descriptors~\cite{Galvez12tro-dbow}.
We augment the appearance descriptor with an object-based descriptor and a place-based descriptor (computed from the objects and places within a radius from the agent node).
The former is computed as the histogram of the object labels in the node's surroundings, intuitively describing the set of nearby objects.
The latter is computed as the histogram of the distances associated to each place in the node surroundings,
intuitively describing the geometry of the map near the node.
While computing the descriptors, we also keep track of the IDs of the objects and places in the agent node's surroundings, which are used for geometric verification.

For loop closure detection we compare the hierarchical descriptor of the current (query) node with all the past agent node descriptors, searching for a match.
When performing loop
closure detection, we walk down the hierarchy of descriptors (from places, to objects, to appearance descriptors).
In particular, when comparing the descriptors of two nodes, we compare the places descriptor and
--if the descriptor distance is below a threshold-- we move on to comparing object descriptors and then appearance descriptors.
If any of the descriptor comparisons return a putative match, we  perform geometric verification;
see Fig.~\ref{fig:lcd} for a visual summary.

\myParagraph{Bottom-up Geometric Verification}
After we have a putative loop closure between our query and match agent nodes (say $i$ and $j$), we
attempt to compute a relative pose between the two by performing a bottom-up geometric verification.
In particular, whenever we have a match at a given layer (\eg between appearance descriptors at the agent layer, or between object descriptors at the object layer), 
we attempt to register frames $i$ and $j$.
For registering visual features we use standard RANSAC-based geometric verification as in~\cite{Rosinol20icra-Kimera}. 
If that fails, we attempt registering
 objects using TEASER++~\cite{Yang20tro-teaser}, discarding loop closures that also fail object registration.
This bottom-up approach has the advantage that putative matches that fail appearance-based geometric verification (\eg due viewpoint or illumination changes) can successfully lead to valid loop closures during the object-based geometric verification.
Section~\ref{sec:experiments} indeed shows that the proposes hierarchical descriptors improve the quality and quantity of detected loop closures.
\subsection{3D Scene Graph Optimization}
\label{sec:SGO}

In order to correct the 3D scene graph in response to a loop closure,
the \emph{Scene Graph Frontend}  ``assembles'' the outputs of the modules described in Section~\ref{sec:incrementalLayers}
into a single 3D scene graph, and then the \emph{Scene Graph Backend}
(i)~optimizes the
graph using a deformation graph approach and
(ii) post-processes the results to
  remove redundant subgraphs corresponding to the robot visiting the same
location multiple times. %

\myParagraph{Scene Graph Frontend}
The frontend builds an initial estimate of the 3D scene graph that is uncorrected for drift.
More precisely,
the frontend takes as input the result of the modules described in Section~\ref{sec:incrementalLayers}): the latest mesh, places subgraph, objects, and pose graph of the agent (all windowed to a radius around the current robot pose).
The corresponding nodes and edges are incrementally added
to the 3D scene graph data structure (which stores the entire scene graph up to the current time).
Then, the frontend populates inter-layer edges from each object or agent node to the nearest place node
in the active window using nanoflann~\cite{Blanco14-nanoflann}.
Finally, the frontend computes a subsampled version of the mesh
that will be optimized in the deformation graph approach described below.
The subsampled mesh is computed via a octree-based vertex clustering mesh simplification approach,
resulting in a smaller subset of nodes (which we refer to as the \emph{mesh control points}) and edges representing connectivity between nodes.

\begin{figure}
    \centering
    \subfloat[Loop closure detection] {
        \centering
        \includegraphics[trim={80, 50, 80, 40}, clip, width=0.47\columnwidth]{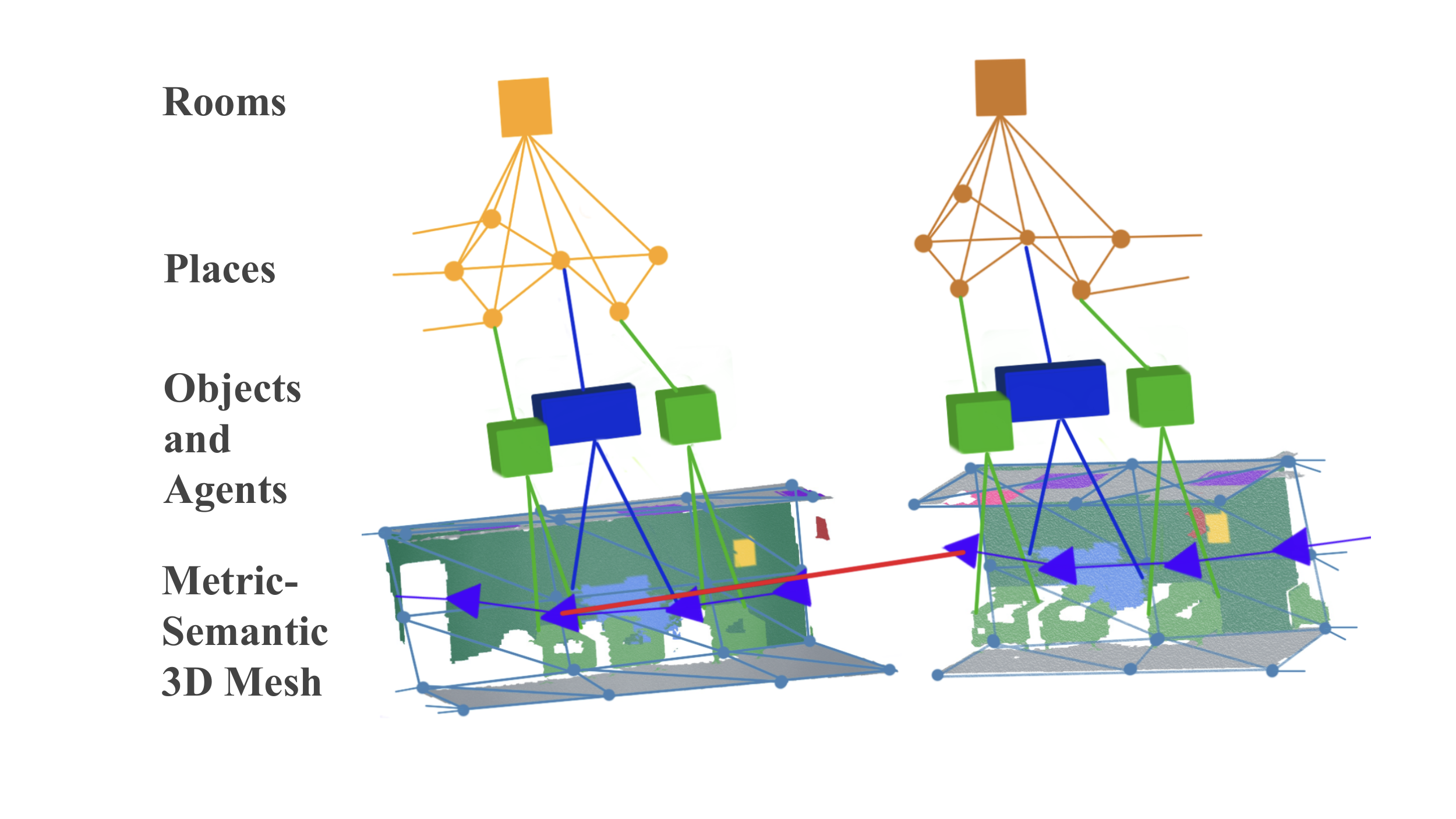}
    }
    \subfloat[Optimization] {
        \centering
        \includegraphics[trim={10, 60, 60, 50}, clip, width=0.47\columnwidth]{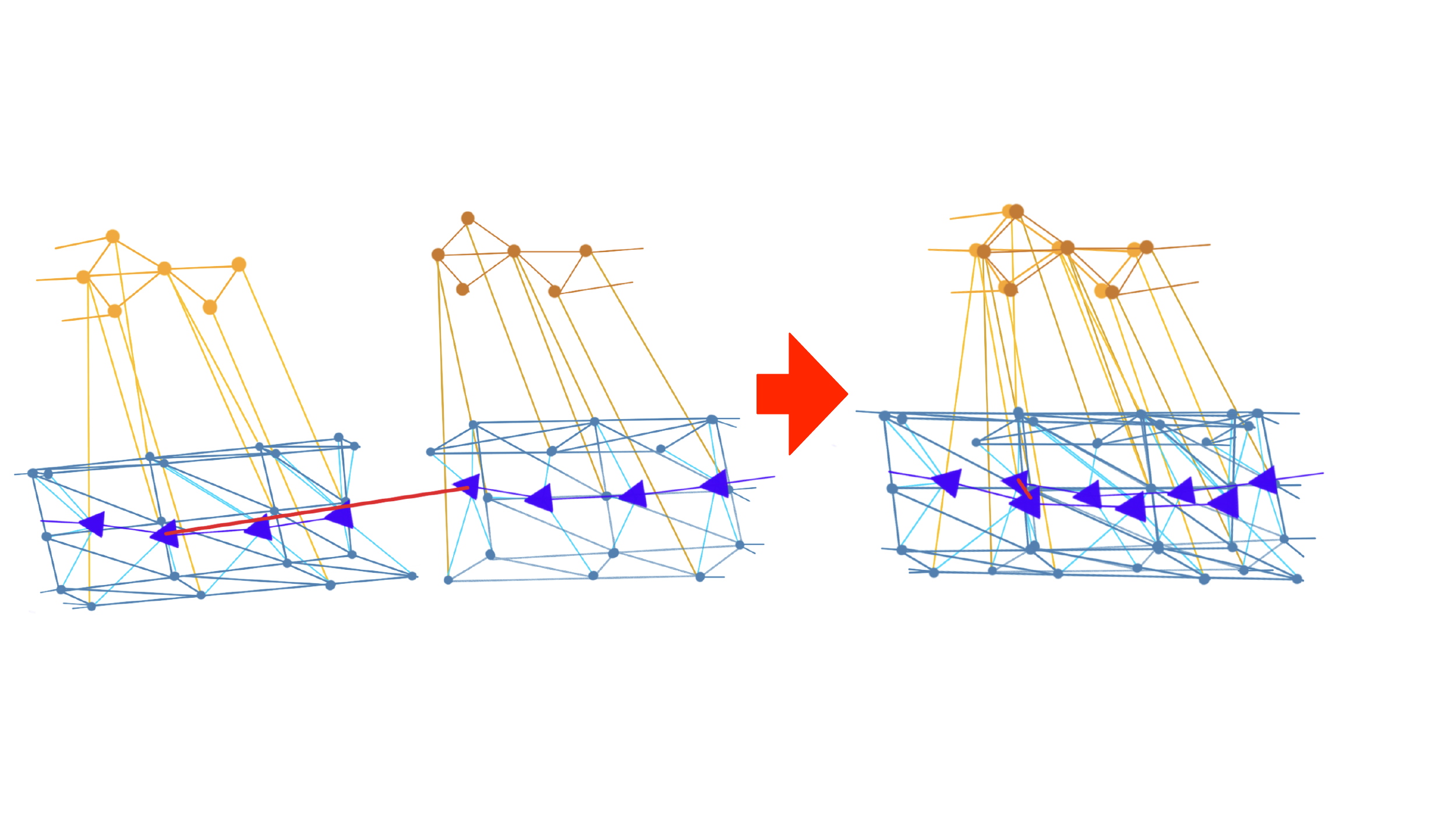}
    }
    \vspace{-2mm}
    \subfloat[Interpolation] {
        \centering
        \includegraphics[trim={210, 50, 150, 80}, clip, width=0.47\columnwidth]{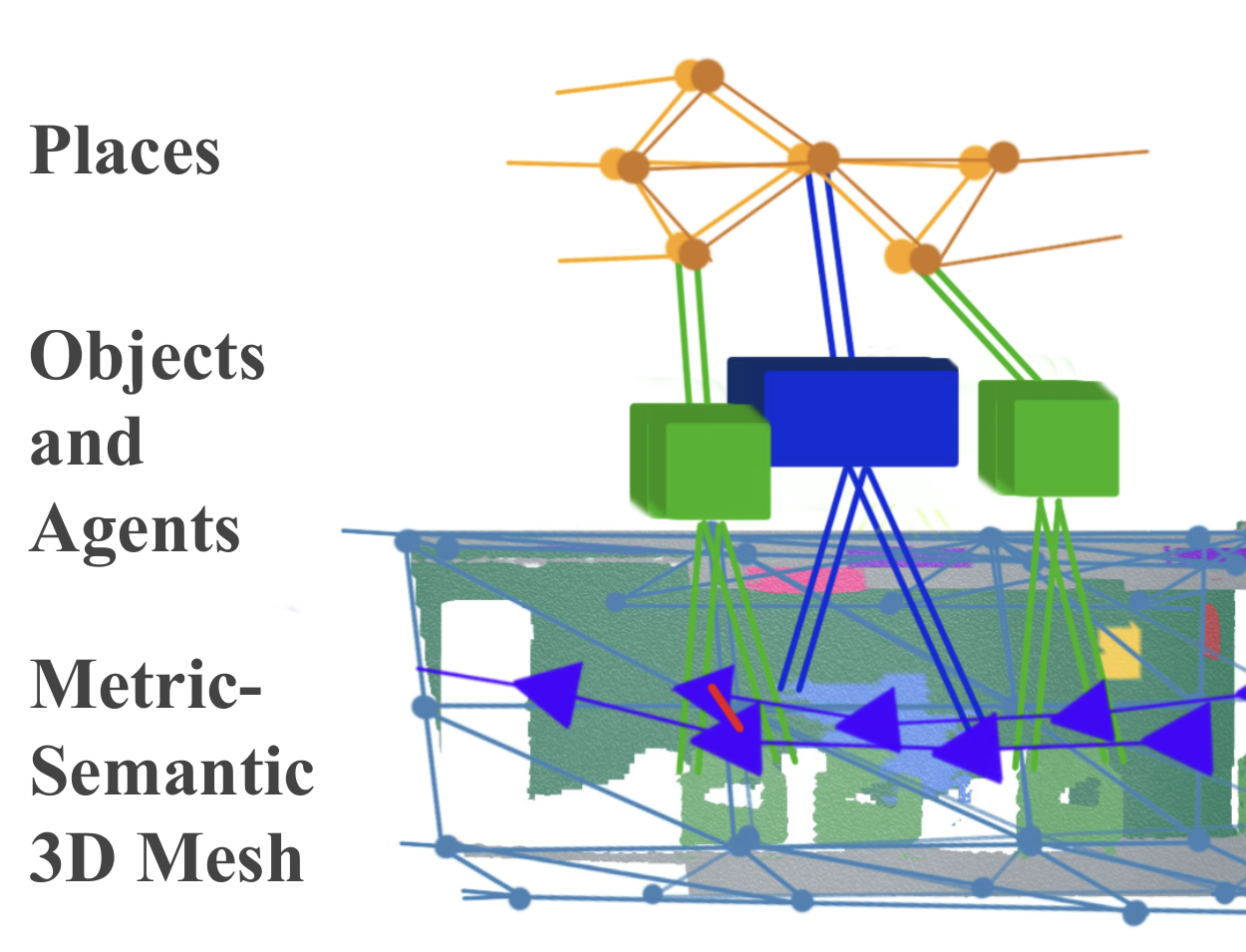}
    }
    \subfloat[Reconciliation] {
        \centering
        \includegraphics[trim={160, 50, 150, 35}, clip, width=0.47\columnwidth]{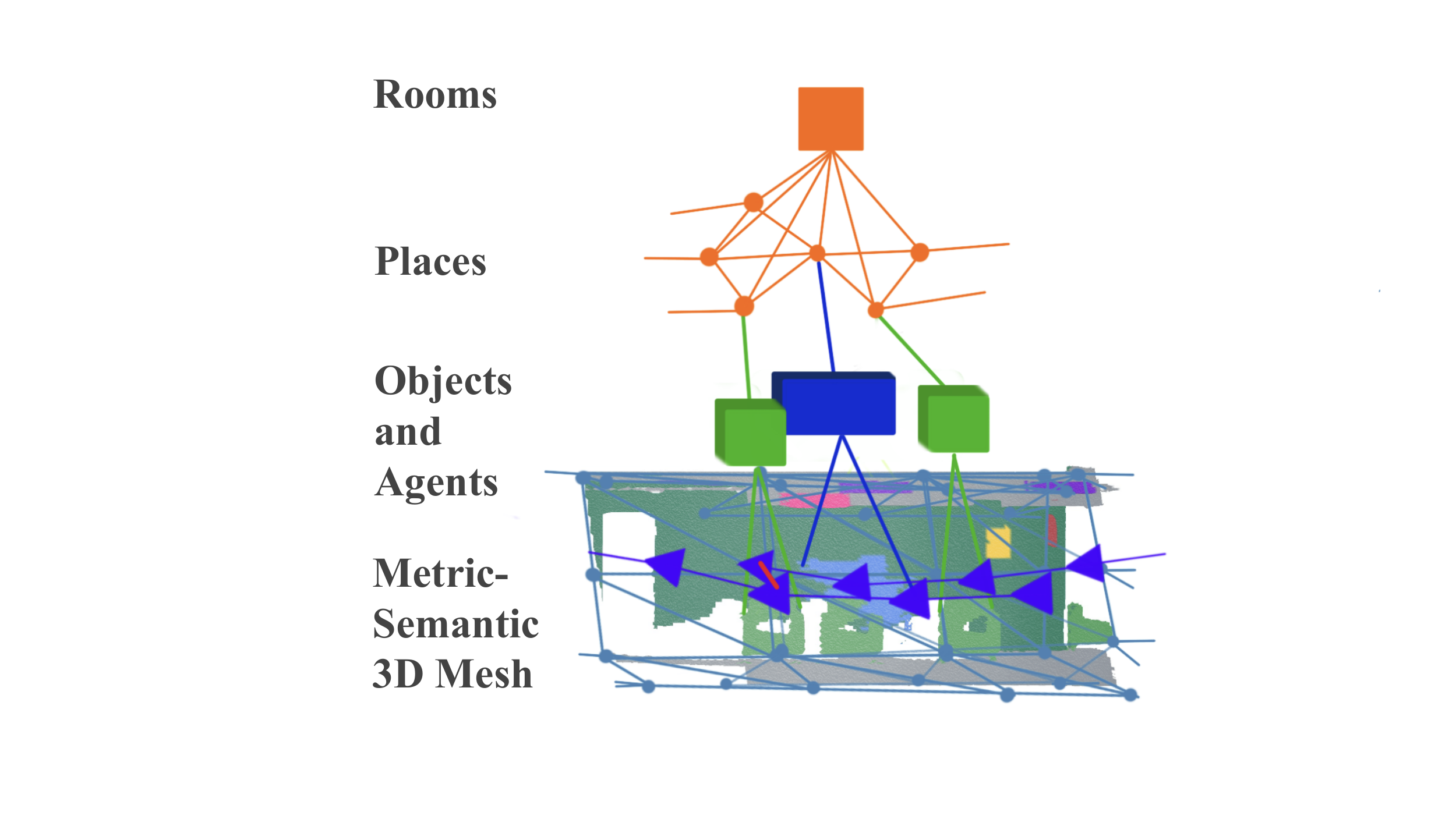}
    }
    \vspace{-1mm}
    \caption{Loop closure detection and optimization:
    (a) after a loop closure is detected, 
    (b) we extract and optimize a subgraph of the 3D scene graph --the \emph{deformation graph}-- that
    includes the agent poses, the places, and a subset of the mesh vertices.
    (c) We then we reconstruct the rest of the graph via interpolation as in~\cite{Sumner07siggraph-embeddedDeformation},
    and (d) reconcile overlapping nodes.
    \vspace{-15mm}
    \label{fig:deformation}}
\end{figure}

\myParagraph{Scene Graph Backend}
When a loop closure is detected, the backend optimizes an \emph{embedded deformation graph}~\cite{Sumner07siggraph-embeddedDeformation}
built from the frontend scene graph and then reconstructs the other nodes in the scene graph via interpolation
as in~\cite{Sumner07siggraph-embeddedDeformation} (Fig.~\ref{fig:deformation}).
More precisely, we form the deformation graph as the subgraph of the 3D scene graph that includes
(i) the agent layer, consisting of a pose graph that includes both odometry and loop closures
edges, (ii) the mesh control points and the corresponding edges, and (iii) the minimum spanning tree of the places layer.
 By construction, these layers form a connected subgraph via the inter-layer edges added by the frontend.
The choice of using the minimum spanning tree of places is mostly motivated by computational reasons:
 the use of the spanning tree preserves the sparsity of the graph.

\begin{figure*}[!t]
    \centering
    \includegraphics[width=0.99\textwidth]{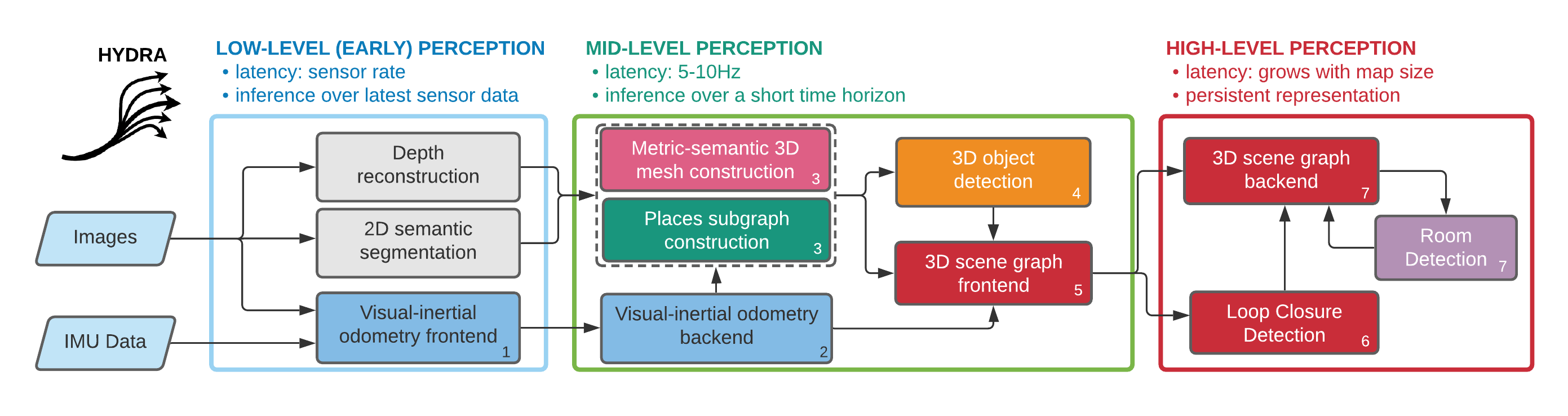}\vspace{-5mm}
    \caption{\name's functional blocks. %
             We conceptualize three different functional block groupings: low-level perception, mid-level perception, and high-level perception in order of increasing latency.
            Each functional block is labeled with a number that identifies the
            \tc{``logical'' thread that the module belongs to.}
             }\vspace{-5mm}\label{fig:architecture}
\end{figure*}

The embedded deformation graph approach associates a local frame (\ie a pose) to each node in the deformation graph and then
solves an optimization problem to adjust the local frames in a way that minimizes
 deformations associated to each edge (including loop closures).
In hindsight, this step transforms a subset of the 3D scene graph into a
\emph{factor graph}~\cite{Cadena16tro-SLAMsurvey}, where edge potentials need to be minimized.
We refer the reader to~\cite{Sumner07siggraph-embeddedDeformation} for the details about the optimization
and note that we use the reformulation of the deformation graph with rigid transforms from~\cite{Rosinol21ijrr-Kimera} (instead of affine as in~\cite{Sumner07siggraph-embeddedDeformation}) to obtain a standard pose graph optimization problem that is amenable to off-the-shelf solvers.
In particular, we use the Graduated Non-Convexity (GNC) solver in GTSAM~\cite{Antonante21tro-outlierRobustEstimation}, which is also able to reject incorrect loop closures as outliers.

Once the optimization is finished,
the place nodes are updated with
their new positions and the full mesh is interpolated based on the deformation graph approach in~\cite{Sumner07siggraph-embeddedDeformation}.
We then recompute the object centroids and bounding boxes from the position of the corresponding vertices in the
newly deformed mesh.
During this update, overlapping nodes are also merged:
for places nodes, we merge nodes within
a distance threshold (0.4m in our implementation);
for object nodes we merge nodes if the corresponding objects have the same semantic label and
 if one of nodes is contained inside the bounding box of the other node.
We maintain a version of the scene graph where the nodes are not merged; this
enables undoing wrong loop closures if an accepted loop closure is deemed to be
an outlier by GNC later on.
Finally, we re-detect rooms from the merged places using the approach described in Section~\ref{sec:incrementalLayers}.

\section{Thinking Fast and Slow: \\ the \name Architecture} %

We implement our \lc{\SPIN} into a highly parallelized architecture, named \emph{\name}.
\name involves a combination of
processes that run at sensor rate (\eg feature tracking for visual-inertial odometry),
at sub-second rate (\eg mesh and place reconstruction),
and at slower rates (\eg the scene graph optimization, whose complexity depends on the map size).
Therefore these processes have to be organized such that slow-but-infrequent computation (\eg scene graph optimization)
does not get in the way of faster processes.

We visualize \name in Fig~\ref{fig:architecture}.
Each block in the figure denotes an algorithmic module matching the discussion in the previous sections.
\name starts with fast \emph{early} perception processes (Fig~\ref{fig:architecture}, left),
which perform low-level perception tasks such as feature detection and tracking (at frame-rate),
2D semantic segmentation, and stereo-depth reconstruction (at keyframe rate).
The result of early perception processes are passed to mid-level perception processes
(Fig~\ref{fig:architecture}, center).
These include algorithms that incrementally construct (an odometric version of) the agent layer
(\eg the visual-inertial odometry backend), the mesh and places layers, and the object layer.
Mid-level perception also includes the scene graph frontend, which collects the result
of the other modules into an ``unoptimized'' scene graph.
Finally, the high-level perception processes perform loop closure detection, 
execute scene graph backend optimization, and perform
room detection.\footnote{While room detection can be performed quickly, it still operates
on the entire graph, hence it is more suitable as a slow high-level perception process.}
This results in a globally consistent, persistent 3D scene graph.

\name runs in real-time on a multi-core CPU; the only module that relies on GPU computing is the
2D semantic segmentation, which uses a standard off-the-shelf deep network.
Running on CPU has the advantage of (i) leaving the GPU to learning-oriented components, and
(ii) being compatible with the power limitations imposed by current mobile robots.

\section{Experiments}
\label{sec:experiments}

This section shows that \name %
builds 3D scene graphs in real-time with an accuracy comparable to batch offline methods.

\subsection{Experimental Setup}

{\bf Datasets.} We utilize two datasets for our experiments: uHumans2 (uH2)~\cite{Rosinol21ijrr-Kimera} and SidPac.
The uH2 dataset is a Unity-based simulated dataset~\cite{Rosinol21ijrr-Kimera} that includes
 three scenes: a
small apartment, an office, and a subway station.
 The dataset provides visual-inertial data as well as ground-truth depth and 2D semantic segmentation. 
The dataset also provides ground truth robot trajectories that we use for benchmarking purposes.

The SidPac dataset is a real dataset collected in a graduate student housing building
 using a visual-inertial hand-held device.
We used a Kinect Azure camera as the primary collection device with an Intel RealSense T265
rigidly attached to the Kinect to provide external odometry input.
The dataset consists of two separate recordings.
The first recording covers two floors of the building (Floors 1 \& 3),
where we walked through a common room, a music room, and a recreation room on the first floor of the graduate residence,
went up a stairwell, through a long corridor as well as a student apartment on the third floor,
then finally down another stairwell to revisit the music room and the common room,
ending where we started.
The second recording also covers two floors (Floors 3 \& 4), where we map student apartments as well as lounge
and kitchen areas which are duplicated across both floors.
These scenes are particularly challenging given the scale of the scenes
(average traversal of around 400 meters), the prevalence of
glass and strong sunlight in regions of the scenes (causing partial
depth estimates from the Kinect), and feature-poor regions in hallways.
We obtain a proxy for the ground-truth trajectory for both SidPac datasets via a hand-tuned pose graph optimization with additional height priors, to reduce drift and qualitatively match the building floor plans. 

{\bf \name.} 
For the real datasets, we use the depth reconstruction from the Kinect (\cf~Fig.~\ref{fig:architecture}) and 
we use HRNet~\cite{Wang21pami-hrnet} for 2D semantic segmentation, 
using the pre-trained model from the MIT Scene Parsing
challenge~\cite{Zhou17cvpr-ade20k}. While newer \tc{and more} performant
networks exist (\eg~\cite{Bao21arxiv-beit, Huang21iccv-fapn,
Dong21arxiv-cswin}), few had pre-trained semantic segmentation models
for ADE20k~\cite{Zhou17cvpr-ade20k} and were compatible
with our inference toolchain (ONNX and TensorRT).
For the simulated datasets, we use the provided depth and segmentations.
For both simulated and real datasets we use Kimera-VIO~\cite{Rosinol20icra-Kimera} for visual-inertial odometry, 
and in the real scenes we fuse the Kimera-VIO estimates with the output of the RealSense T265 to improve the 
quality of the odometric trajectory. 
All the remaining blocks in~Fig.~\ref{fig:architecture} are implemented in C++, following the approach described in this paper.
In the experiments we use a workstation with an AMD Ryzen9 3960X with 24 cores and two Nvidia GTX3080s, \lc{though we also report timing results on an embedded computer (Nvidia Xavier NX) at the end of this section}.

\subsection{Results and Ablation Study}

We present an extensive evaluation of the accuracy and runtime of our real-time approach against the batch offline scene graph construction approach from~\cite{Rosinol21ijrr-Kimera}. 

\begin{figure*}
    \centering
    \includegraphics[trim={0mm 0mm 0mm 0mm}, clip, width=\textwidth]{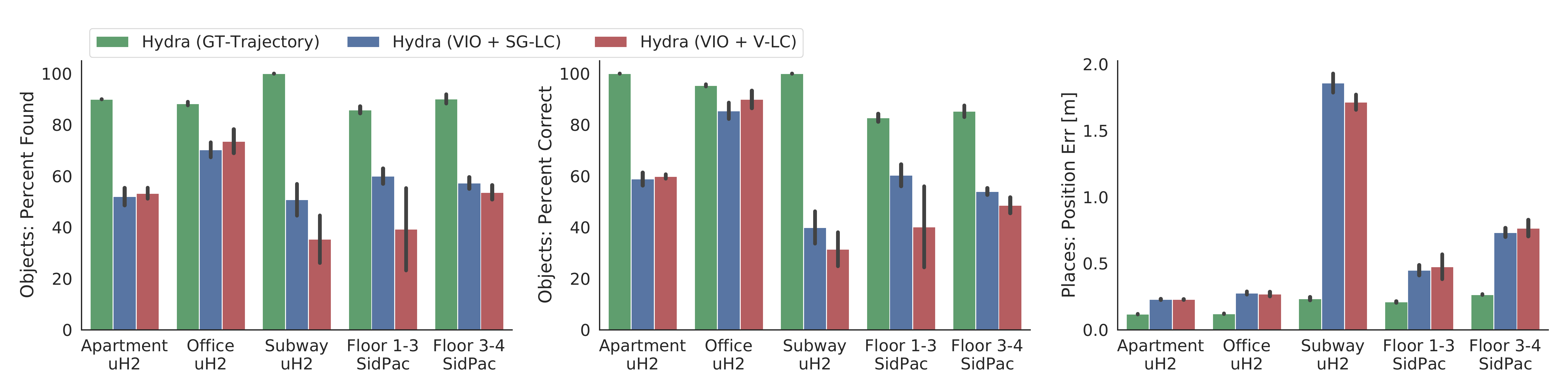}\vspace{-2mm}
    \caption{Accuracy of the objects and places estimated by \name.
             Each plot reports the mean across 5 trials along with the standard deviation as an error bar.
             }\vspace{-6mm}\label{fig:op_metrics}
\end{figure*}

\myParagraph{Accuracy Evaluation: Objects and Places} 
Fig.~\ref{fig:op_metrics} evaluates the object and place layers by comparing three different configurations of \name.
The first configuration (``\igx'') uses ground-truth poses to incrementally construct the scene graph.
The second and third configurations (``\ivl'' and ``\ivd'' respectively) use visual-inertial odometry (VIO) for odometry estimation 
and then use vision-based loop closures (\ivl) or the proposed scene graph loop closures (\ivd).
For the objects and places evaluation, we consider the batch scene graph constructed from ground-truth poses using~\cite{Rosinol21ijrr-Kimera}   as the ground-truth scene graph.

\begin{figure}
    \centering
    \includegraphics[trim={0mm 0mm 0mm 0mm}, clip, width=\columnwidth]{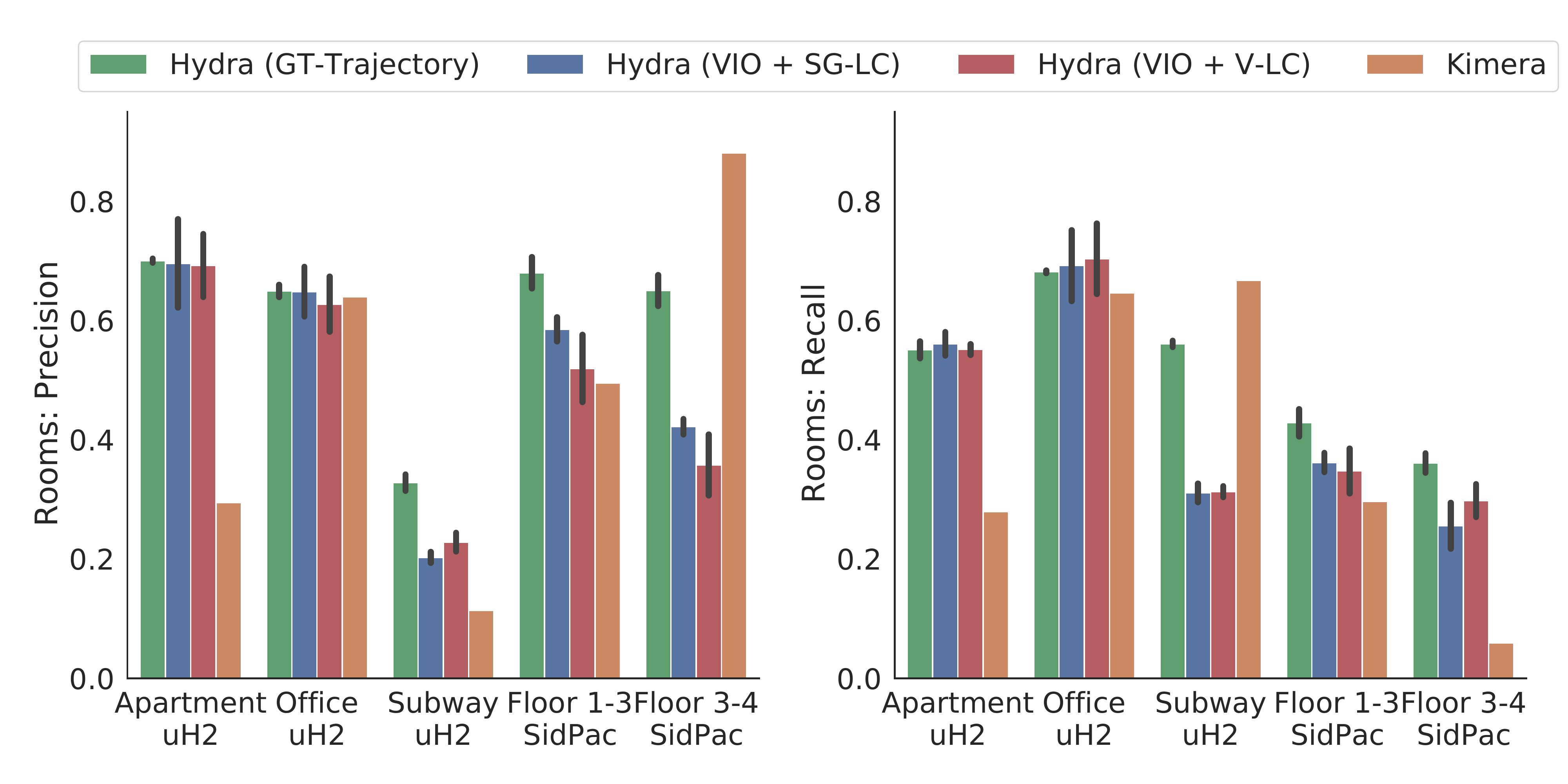}\vspace{-2mm}
    \caption{Room detection accuracy of \name versus Kimera~\cite{Rosinol21ijrr-Kimera}.
             Each plot reports the mean across 5 trials along with the standard deviation as an error bar except Kimera, for which one trial is shown (due to its slower runtime).
             }\vspace{-5mm}\label{fig:room_metrics}
\end{figure}
 For the object layer, we report two metrics:
the percentage of objects in the ground-truth scene graph that have an estimated object with the correct semantic label within a specified radius (``\percFound'') and
the percentage of objects in the estimated scene graph that have a ground-truth object with the correct semantic label  within a specified radius (``\percCorrect'').\footnote{As distance thresholds, we use 0.2m for the apartment scene, 0.3m for the office scene, 4m for the subway scene, 0.9m for floors 1\&3 of SidPac and 1.4m for floors 3\&4 of SidPac.
These thresholds were chosen to roughly correspond to the mean Absolute Trajectory Error (ATE) for each scene, in 
order to normalize the metrics according to the environment size.}
For the places layer, we measure the mean distance of an estimated place node to the nearest voxel in the ground-truth 
GVD (``\positionError'').

We note some important trends in Fig.~\ref{fig:op_metrics}.
First, when using \igx, \name's performance is close to the ground-truth scene graph (80-100\% found and correct objects, sub-25cm places position error). This demonstrates that --given the trajectory-- the real-time scene graph from \name is \tc{comparable}%
 to the batch and offline approaches at the state of the art.
Second, \ivl and \ivd maintain reasonable levels of accuracy for the objects and places 
and attain comparable performance in small to medium-sized scenes (\eg Apartment, Office).  %
In these scenes, the drift is small and the loop closure strategy does not radically impact performance (differences are within standard deviation, shown as black confidence bars). 
However, in larger scenes (\eg SidPac) loop closures are more important and  \ivd substantially outperforms \ivl in terms of object accuracy. Importantly, using \ivd typically leads to a reduction in the standard deviation of the results, confirming that the proposed approach for scene graph loop closure detection leads to more reliable loop closure results (more details and ablations below). The place positions errors remain similar for both \ivd and \ivl
and are larger in the subway dataset, which includes larger open-spaces 
with more distant nodes in the subgraph of places.
\myParagraph{Accuracy Evaluation: Rooms} 
Fig.~\ref{fig:room_metrics} evaluates the room detection performance, using 
the precision and recall metrics defined in~\cite{Bormann16icra-roomSegmentationSurvey} (here
we compute precision and recall over 3D voxels instead of 2D pixels). \lc{More formally, these metrics are:}
\begin{equation}\label{eq:room_pr}
\begin{split}
\text{\emph{Precision}} &= \frac{1}{|R_e|}\sum_{r_e \in R_e} \max_{r_g \in R_g} \frac{|r_g \cap r_e|}{|r_e|} \\
\text{\emph{Recall}} &= \frac{1}{|R_g|}\sum_{r_g \in R_g} \max_{r_e \in R_e} \frac{|r_e \cap r_g|}{|r_g|}
\end{split}
\end{equation}
\lc{where $R_e$ is the set of estimated rooms, $R_g$ is the set of ground-truth rooms, and $|\cdot|$ returns the cardinality of a set;  here, each room $r_e$ (or $r_g$) is defined as a set of free-space voxels.}
\lc{We hand-label the {ground-truth} rooms $R_g$ from the ground-truth reconstruction of the environment. %
In particular, we manually define \tc{sets of} bounding boxes for each room and
assign a unique (ground-truth) label to the free-space voxels falling within each room. 
For the estimated rooms $R_e$, we derive the free-space voxels from the places comprising each estimated room.}
In~eq.~\eqref{eq:room_pr}, \tc{\emph{Precision} then measures the maximum overlap in voxels with a ground-truth room for every estimated room, and \emph{Recall} measures the maximum overlap in voxels with an estimated room for every ground-truth room.}
Intuitively, low precision corresponds to under-segmentation, \ie fewer and larger room estimates, and low recall corresponds to over-segmentation, \ie more and smaller room estimates.
\lc{For benchmarking purposes, we also include the approach in~\cite{Rosinol21ijrr-Kimera} (\emph{Kimera}) as a baseline for evaluation.} 

Fig.~\ref{fig:room_metrics} shows that while Kimera~\cite{Rosinol21ijrr-Kimera} estimates a room segmentation with similar precision and recall to \name (\igx) for the Office scene (the only single-floor scene), \name excels in multi-floor environments.
For the split-level Apartment scene, we achieve significantly \lc{higher} precision and recall as compared to Kimera.
For SidPac Floor 3-4, the difference is particularly dramatic, where the Kimera achieves 0.88 precision but only 0.06 recall, as it only is able to segment 2 of 10 ground-truth rooms. In general, our approach estimates a room segmentation with consistent precision and recall (if slightly over-segmented), while Kimera oscillates between either low-precision and high-recall estimates (\ie extreme under-segmentation), or high-precision and low-recall estimates (\ie where it fails to segment most of the rooms).
These differences stem from the difficulty of setting an appropriate height to attempt to segment rooms at for Kimera.
Finally, it is worth noting that our room segmentation approach is able to mostly maintain the same levels of precision and recall for \ivd and \ivl despite drift.

\begin{figure}
    \centering
    \includegraphics[trim={0mm 0 0mm 0}, clip, width=0.95\columnwidth]{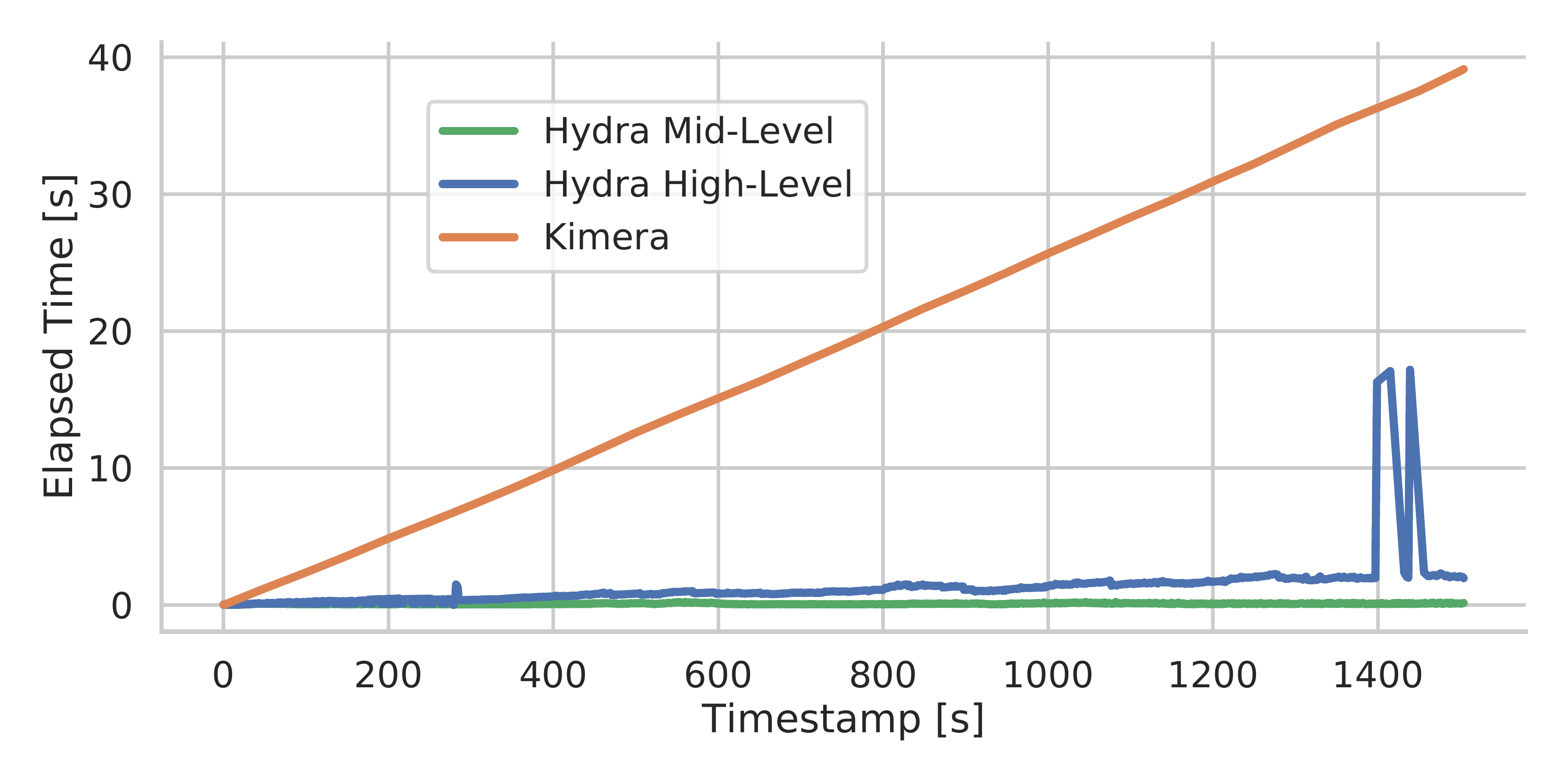}\vspace{-3mm}
    \caption{Runtime required for scene graph construction vs. timestamp for the SidPac Floor 1-3 dataset for a batch approach (Kimera) and for the proposed incremental approach (\name).
         For timing of the low-level processes in \name, we refer the reader to the analysis in~\cite{Rosinol21ijrr-Kimera}, as we also rely on Kimera-VIO.
    }\vspace{-2mm}\label{fig:timing}
\end{figure}
\setlength{\tabcolsep}{5pt}
\begin{table}[t!]
    \centering
    \begin{tabular}{cl ccc}
        \toprule
        & & \multicolumn{3}{c}{Layer} \\
        \cmidrule(l{2pt}r{2pt}){3-5}
        & & Objects [ms] & Places [ms] & Rooms [ms] \\
        \midrule

        \multirow{3}{*}{uH2}          & Apartment    & $32.4 \pm 12.9$ & $5.3 \pm 1.4$ & $4.4 \pm 2.1$ \\
                                      & Office       & $24.1 \pm 12.8$ & $8.1 \pm 1.3$ & $19.0 \pm 12.3$ \\
                                      & Subway       & $9.8 \pm 9.3$ & $5.9 \pm 0.7$ & $16.5 \pm 10.6$ \\
        \midrule
        \multirow{2}{*}{SP}           & Floor 1-3    & $50.4 \pm 30.3$ & $3.4 \pm 1.0$ & $11.4 \pm 14.4$ \\
                                      & Floor 3-4    & $75.3 \pm 37.0$ & $4.2 \pm 2.1$ & $15.0 \pm 14.6$ \\

        \bottomrule
    \end{tabular}
    \caption{\name: timing breakdown}\vspace{-7mm}\label{tab:component_timing}
\end{table}
 
\myParagraph{Runtime Evaluation} 
Fig.~\ref{fig:timing} reports
the runtime of \name versus the batch approach in~\cite{Rosinol21ijrr-Kimera}.
This plot shows that the runtime of the batch approach increases over time and takes more than
40 seconds to generate the entire scene graph for moderate scene sizes;
as we mentioned, most processes in the batch approach~\cite{Rosinol21ijrr-Kimera} entail processing the entire \ESDF (\eg place extraction and room detection), inducing a linear increase in the runtime as the \ESDF grows.
On the other hand, our scene graph frontend (\emph{\name Mid-Level} in Fig.~\ref{fig:timing}) has a fixed computation cost.
In Fig.~\ref{fig:timing}, a slight upward trend is observable for \emph{\name High-Level}, driven by room detection and scene graph optimization computation costs, though remaining much lower than batch processing.
Noticeable spikes in the runtime for \emph{\name High-Level} (\eg at 1400 seconds) correspond to the execution of the 3D scene graph optimization when new loop closures are added.

Table~\ref{tab:component_timing} reports timing breakdown for the incremental creation of each layer across scenes for a single trial.
The object layer runtime is determined by the number of mesh vertices in the active window with object semantic class;
hence the SidPac scenes have a higher computation cost than the other scenes.
The room layer runtime is determined by the number of places (a combination of how complicated and large the scene is);
this is why the Office has the largest computation cost for the rooms despite being smaller than the SidPac scenes.

\lc{
While the timing results in Table~\ref{tab:component_timing} are obtained with a relatively powerful workstation,
here we restate that \name can run in real-time on embedded computers commonly used in robotics applications. 
Towards this goal, we also measure timing statistics on an Nvidia Xavier NX for the uHumans2 Office scene.
\name processes the objects in $75 \pm 35$~ms, the places in $33 \pm 6$~ms and rooms in $55 \pm 41$~ms.
Note that our target ``real-time'' rate for these layers is keyframe rate (5 Hz). 
While there is still margin to optimize computation (see conclusions), 
these initial results stress the practicality and real-time capability of \name in building 3D scene graphs.}
\begin{figure}
    \centering
    \includegraphics[trim={30mm 0 30mm 0}, clip, width=\columnwidth]{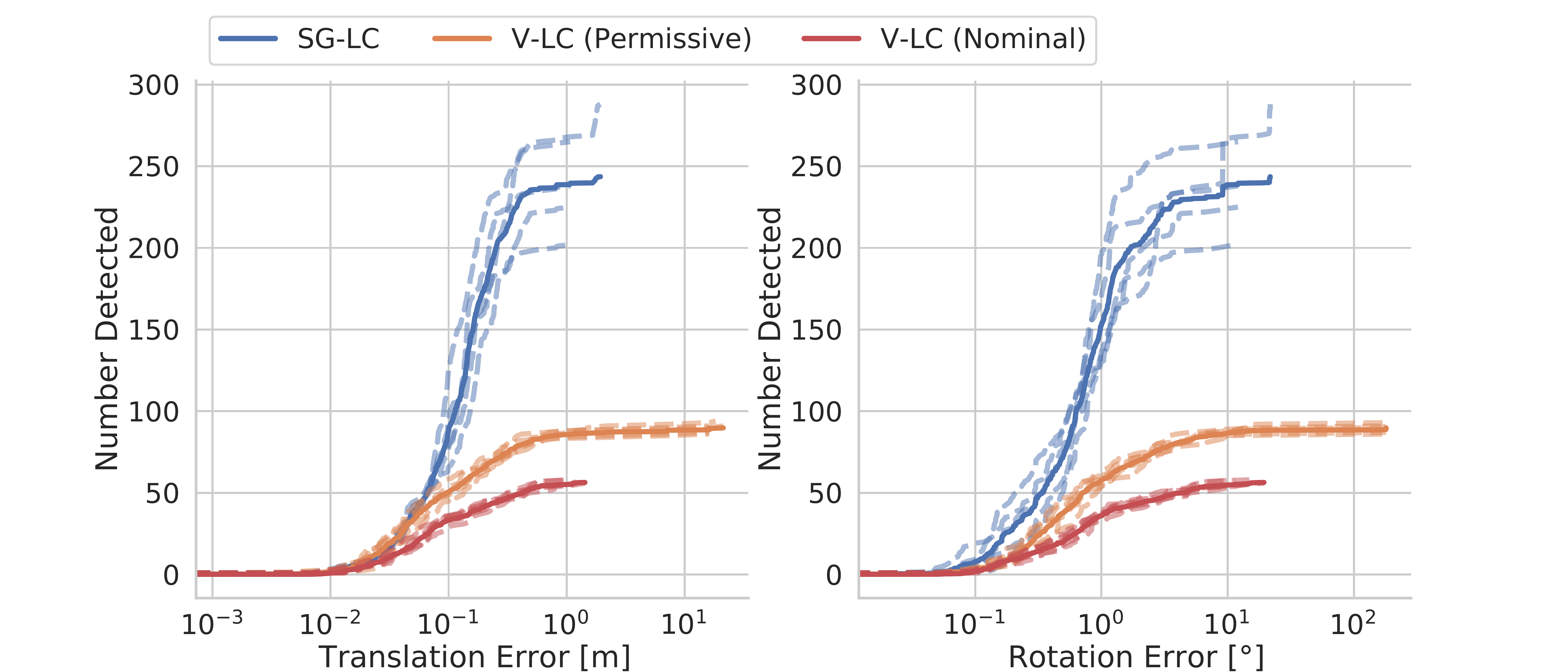}\vspace{-2mm}
    \caption{Number of detected loop closures versus error of the estimated loop closure pose
             for three different loop closure detection configurations. Five individual trials and a trend-line are shown for each configuration.}\vspace{-5mm}\label{fig:lcd_experiment}
\end{figure}
 
\myParagraph{Loop Closure Ablation Study} Finally, we take a closer look at the quality of the loop closures candidates proposed by our hierarchical loop closure detection
approach, and compare it against traditional vision-based approaches on the Office scene. %
In particular, we compare our approach against a 
vision-based loop closure detection that uses DBoW2 for place recognition 
and ORB feature matching, as described in~\cite{Rosinol21ijrr-Kimera}.

Fig.~\ref{fig:lcd_experiment} shows the number of detected loop closures against the error of the registered solution (\ie the relative pose between query and match computed by the geometric verification) for three different loop closure configurations:
(i) ``SG-LC'': the proposed scene graph loop closure detection, %
(ii) ``V-LC (Nominal)'': a traditional vision-based loop closure detection with nominal parameters (same as~Fig.~\ref{fig:op_metrics}), %
and (iii) ``V-LC (Permissive)'': a vision-based loop closure detection with more permissive parameters (\ie a decreased score threshold and less restrictive geometric verification settings). 
{We report key parameters used in this evaluation in the appendix.}
As expected, making the vision-based detection parameters more permissive leads to more but lower-quality loop closures.
On the other hand, the scene graph loop closure approach produces approximately twice as many loop closures within 10cm of error and 1 degree of error as the permissive vision-based approach.
The proposed approach produces quantitatively and quantitatively better loop closures compared to both baselines. 

\section{Conclusions} %
\label{sec:conclusions}

This paper introduces \name, 
\tc{a} \emph{real-time \lc{\SPIN}} that builds 
a 3D scene graph from sensor data in real-time. 
\name runs at sensor rate thanks to the combination of novel online algorithms and a highly parallelized 
perception architecture.
Moreover, it allows building a \emph{persistent} representation of the environment thanks to a novel approach for 3D scene graph optimization. 
While we believe the proposed approach constitutes a substantial step towards high-level 3D scene understanding for robotics, 
\name can be improved in many directions.
First, some nodes in the reconstructed 3D scene graph are unlabeled (\eg the algorithms in this paper are able to detect rooms, but are unable to label a given room as a ``kitchen'' or a ``bedroom''); future work includes bridging \name with learning-based methods for 3D scene graph node labeling~\cite{Talak21neurips-neuralTree}. 
Second, it would be interesting to label nodes and edges of the 3D scene graph with a richer set of relations and affordances, building on~\cite{Wu21cvpr-SceneGraphFusion}.
\lc{Third, the connections between our scene graph optimization approach and pose graph optimization offer opportunities to improve the efficiency of the optimization by leveraging recent advances in pose graph sparsification.}
Finally, the implications of using 3D scene graphs for prediction, planning, and decision-making are mostly unexplored (see~\cite{Ravichandran21icra-RLwithSceneGraphs,Agia21corl-taskography} for early examples), which opens further avenues for future work.

\section*{Disclaimer}

Research was sponsored by the United States Air Force Research Laboratory and the United States Air Force Artificial Intelligence Accelerator and was accomplished under Cooperative Agreement Number FA8750-19-2-1000. The views and conclusions contained in this document are those of the authors and should not be interpreted as representing the official policies, either expressed or implied, of the United States Air Force or the U.S. Government. The U.S. Government is authorized to reproduce and distribute reprints for Government purposes notwithstanding any copyright notation herein.
 
\section*{Acknowledgments}

This work was partially funded by the AIA CRA FA8750-19-2-1000, ARL DCIST CRA W911NF-17-2-0181, and ONR RAIDER N00014-18-1-2828.

\bibliographystyle{plainnat}

\appendix

\myParagraph{Loop Closure Ablation Parameters}
Table~\ref{table:params-V} reports key parameters used for 
``V-LC (Permissive)'' and ``V-LC (Nominal)'' 
in  the loop closure ablation study in Fig.~\ref{fig:lcd_experiment}. 
For the meaning of each parameter we refer the reader to the open-source 
Kimera implementation released by~\cite{Rosinol21ijrr-Kimera}.

\setlength{\tabcolsep}{4pt}
\begin{table}[h]
    \centering
    \begin{tabular}{ccc}
        \toprule
        Parameter & V-LC (Permissive) & V-LC (Nominal) \\
        \midrule
        L1 Score Threshold & 0.05 & 0.4 \\
        Minimum NSS & 0.005 & 0.05 \\
        Min. RANSAC Correspondences & 12 & 15 \\
        5pt RANSAC Inlier Threshold & 0.01 & 0.001 \\
        Lowe Matching Ratio & 0.8 & 0.9 \\
        \bottomrule
    \end{tabular}
    \caption{Visual loop closure parameters \label{table:params-V}}
\end{table}

Table~\ref{table:params-SG} reports key parameters used for ``SG-LC''
in  the ablation study in Fig.~\ref{fig:lcd_experiment}. 
All scene graph descriptors were computed with a radius of 13 meters.
SG-LC does not use NSS (Normalized Similarity Scoring) to filter out matches. 
\begin{table}[h]
    \centering
    \begin{tabular}{ccc}
        \toprule
        Parameter & SG-LC \\
        \midrule
        Agent L1 Match Threshold & 0.01 \\
        Object L1 Match Threshold & 0.3 \\
        Places L1 Match Threshold & 0.5 \\
        Min. RANSAC Correspondences & 15 \\
        5pt RANSAC Inlier Threshold & 0.001 \\
        Object L1 Registration Threshold & 0.8 \\
        Object Minimum Inliers & 5 \\
        TEASER Noise Bound [m] & 0.1 \\
        \bottomrule
    \end{tabular}
    \caption{Scene graph loop closure parameters \label{table:params-SG}}
\end{table}

\end{document}